\documentclass{article}

\usepackage{amsmath,amsfonts,bm}

\def\eqref#1{equation~\ref{#1}}

\def\1{\bm{1}}

\def\rva{{\mathbf{a}}}

\def\rvo{{\mathbf{o}}}

\def\rvs{{\mathbf{s}}}

\def\rvz{{\mathbf{z}}}

\DeclareMathAlphabet{\mathsfit}{\encodingdefault}{\sfdefault}{m}{sl}
\SetMathAlphabet{\mathsfit}{bold}{\encodingdefault}{\sfdefault}{bx}{n}

\def\gA{{\mathcal{A}}}

\def\gD{{\mathcal{D}}}

\def\gH{{\mathcal{H}}}

\def\gM{{\mathcal{M}}}
\def\gN{{\mathcal{N}}}
\def\gO{{\mathcal{O}}}

\def\gQ{{\mathcal{Q}}}

\def\gS{{\mathcal{S}}}

\newcommand{\E}{\mathbb{E}}

\newcommand{\KL}{D_{\mathrm{KL}}}

\DeclareMathOperator*{\argmax}{arg\,max}

\usepackage{graphicx}
\graphicspath{{./figs/}}
\usepackage{xcolor}
\definecolor{mycolor}{RGB}{0,128,255}
\usepackage[colorlinks=true,allcolors=mycolor,pageanchor=true,plainpages=false,pdfpagelabels,bookmarks,bookmarksnumbered]{hyperref}
\usepackage{multirow}
\usepackage{bm}
\usepackage{svg}
\usepackage[capitalise,nameinlink,noabbrev]{cleveref}
\usepackage{subfig}
\usepackage{wrapfig}
\usepackage[inline]{enumitem}

\usepackage{todonotes}

\usepackage[accepted]{icml2020}
\icmltitlerunning{Improving Sample Efficiency in Model-Free
Reinforcement Learning from Images}

\begin{document}

\twocolumn[
\icmltitle{Improving Sample Efficiency in Model-Free
Reinforcement Learning \\from Images}

\icmlsetsymbol{equal}{*}

\begin{icmlauthorlist}
\icmlauthor{Denis Yarats}{nyu,fair}
\icmlauthor{Amy Zhang}{mcgill,mila,fair}
\icmlauthor{Ilya Kostrikov}{nyu}
\icmlauthor{Brandon Amos}{fair}
\icmlauthor{Joelle Pineau}{mcgill,mila,fair}
\icmlauthor{Rob Fergus}{nyu,fair}

\end{icmlauthorlist}

\icmlaffiliation{nyu}{New York University}
\icmlaffiliation{fair}{Facebook AI Research}
\icmlaffiliation{mcgill}{McGill University}
\icmlaffiliation{mila}{MILA}

\icmlcorrespondingauthor{Denis Yarats}{denisyarats@cs.nyu.edu}

\icmlkeywords{Reinforcement Learning, Representation Learning, Continuous Control}

\vskip 0.3in
]

\printAffiliationsAndNotice{}

\newif\ifincludeappendix

\begin{abstract}
Training an agent to solve control tasks directly from high-dimensional images with model-free reinforcement learning (RL) has proven difficult. A promising approach is to learn a latent representation together with the control policy. However, fitting a high-capacity encoder using a scarce reward signal is sample inefficient and leads to poor performance.
Prior work has shown that auxiliary losses, such as image reconstruction, can aid efficient representation learning.  
However, incorporating reconstruction loss into an off-policy learning algorithm often leads to training instability. We explore the underlying reasons and 
identify variational autoencoders, used by previous investigations, as the cause of the divergence.   
Following these findings, we propose effective techniques to improve training stability. 
This results in a simple approach capable of
matching state-of-the-art model-free and model-based algorithms on MuJoCo control tasks. Furthermore, our approach demonstrates robustness to observational noise, surpassing existing approaches in this setting. Code, results, and videos are anonymously available at~\href{https://sites.google.com/view/sac-ae/home}{https://sites.google.com/view/sac-ae/home}.

\end{abstract}

\section{Introduction}

Cameras are a convenient and inexpensive way to acquire state information, especially in complex, unstructured environments, where effective control requires access to the  proprioceptive state of the underlying dynamics. Thus, having effective RL approaches that can utilize pixels as input would potentially enable solutions for a wide range of real world applications, for example robotics. 

The challenge is to efficiently learn a mapping from pixels to an appropriate representation for control using only a sparse reward signal. Although deep convolutional encoders can learn good representations (upon which a policy can be trained), they require large amounts of training data. As existing reinforcement learning approaches already have poor sample complexity, this makes direct use of pixel-based inputs prohibitively slow. For example, model-free methods on Atari~\citep{bellemare13arcade} and DeepMind Control (DMC)~\citep{tassa2018dmcontrol} take tens of millions of steps~\citep{mnih2013dqn,barth-maron2018d4pg}, which is impractical in many applications, especially robotics.

Some natural solutions to improve sample efficiency are i) to use off-policy methods and ii) add an auxiliary task with an unsupervised objective. Off-policy methods enable more efficient sample re-use, while the simplest auxiliary task is an autoencoder with a pixel reconstruction objective. Prior work has attempted to learn state representations from pixels with autoencoders, utilizing a two-step training procedure, where the representation is first trained via the autoencoder, and then either with a policy learned on top of the fixed representation~\citep{lange10deepaeinrl,munk2016mlddpg,higgins2017darla,zhang2018ddr,nair2018imaginedgoal,dwibedi2018visrepr}, or with planning~\citep{mattner2012ae,finn2015deepspatialae}. This allows for additional stability in optimization by circumventing dueling training objectives but leads to suboptimal policies. Other work utilizes continual model-free learning with an auxiliary reconstruction signal in an on-policy manner~\citep{jaderberg2016unreal,shelhamer16selfsuperrl}. 
However, these methods do not report of learning representations and a policy jointly in the off-policy setting, or note that it performs poorly~\citep{shelhamer16selfsuperrl}.

We revisit the concept of adding an autoencoder to model-free RL approaches, with a focus on \emph{off-policy} algorithms. We perform a sequence of careful experiments to understand why previous approaches did not work well. We confirm that a pixel reconstruction loss is vital for learning a good representation, specifically when trained jointly, but requires careful design choices to succeed. Based on these findings, we recommend a simple and effective autoencoder-based \emph{off-policy} method that can be trained \emph{end-to-end}. We believe this to be the first \emph{model-free} \emph{off-policy} approach to train the latent state representation and policy \emph{jointly} and match performance with state-of-the-art model-based methods \footnote{We define model-based methods as those that train a dynamics model. By this definition, SLAC~\cite{lee2019slac} is a model-based method.}~\citep{hafner2018planet,lee2019slac} on many challenging control tasks. In addition, we demonstrate robustness to observational noise and outperform prior methods in this more practical setup.

This paper makes three main contributions: (i) a methodical study of the issues involved with combining autoencoders with model-free RL in the off-policy setting that advises a successful variant we call \textbf{SAC+AE}; (ii) a demonstration of the robustness of our model-free approach over model-based methods on tasks with noisy observations;  and (iii) an open-source PyTorch implementation of our simple and effective algorithm for researchers and practitioners to build upon.

\section{Related Work}

Efficient learning from high-dimensional pixel observations has been a problem of paramount importance for model-free RL. While some impressive progress has been made applying model-free RL to domains with simple dynamics and discrete action spaces~\citep{mnih2013dqn}, attempts to scale these approaches to complex continuous control environments have largely been unsuccessful, both in simulation and the real world. A glaring issue is that the RL signal is much sparser than in supervised learning, which leads to sample inefficiency, and higher dimensional observation spaces such as pixels worsens this problem. 

One approach to alleviate this problem is by training with auxiliary losses.
Early work~\citep{lange10deepaeinrl} explores using deep autoencoders to learn feature spaces in visual reinforcement learning, crucially \citet{lange10deepaeinrl} propose to recompute features for all collected experiences after each update of the autoencoder, rendering this approach impractical to scale to more complicated domains. Moreover, this method has been only demonstrated on toy problems.
Alternatively, \citet{finn2015deepspatialae} apply deep autoencoder pretraining to real world robots that does not require iterative re-training, improving upon computational complexity of earlier methods. However, in this work the linear policy is trained separately from the autoencoder, which we find to not perform as well as end-to-end methods.

\begin{table*}[t!]

\centering
 \resizebox{0.7\linewidth}{!}{
\begin{tabular}{|l|c|cccc|}
\hline
\multirow{2}{*}{Task name}   & Number of & \multirow{2}{*}{SAC:pixel} & \multirow{2}{*}{PlaNet} & \multirow{2}{*}{SLAC} & \multirow{2}{*}{SAC:state} \\
& Episodes & & & &  \\
\hline
\texttt{finger\_spin} & 1000 & $645 \pm 37$ & $659 \pm 45$ & \bm{$900 \pm 39$} & \bm{$945 \pm 19$}\\
\texttt{walker\_walk} & 1000 & $33 \pm 2$ & $949 \pm 9$ & $864 \pm 35$ & \bm{$974 \pm 1$}\\
\texttt{ball\_in\_cup\_catch} & 2000 & $593 \pm 84$ & $861 \pm 80$ & $932 \pm 14$ & \bm{$981 \pm 1$}\\
\texttt{cartpole\_swingup} & 2000 & $758 \pm 58$ & $802 \pm 19$ & - & \bm{$860 \pm 8$}\\
\texttt{reacher\_easy} & 2500 & $121 \pm 28$ & \bm{$949 \pm 25$} & - & \bm{$953 \pm 11$}\\
\texttt{cheetah\_run} & 3000 & $366 \pm 68$ & $701 \pm 6$ & \bm{$830 \pm 32$} & \bm{$836 \pm 105$}\\
\hline
\end{tabular}}
\caption{ \label{table:mf_baseline}
\small{A comparison  of current methods: SAC from pixels, PlaNet, SLAC, SAC from  proprioceptive states (representing an upper bound). The large performance gap between SAC:pixel and SAC:state motivates us to address the representation learning bottleneck in model-free off-policy RL.}}
\end{table*}

\citet{shelhamer16selfsuperrl} employ auxiliary losses to enhance performance of A3C~\citep{mnih2016a3c} on Atari. They recommend a multi-task setting and learning dynamics and reward to find a good representation, which relies on the assumption that the dynamics in the task are easy to learn and useful for learning a good policy. To prevent instabilities in learning, \citet{shelhamer16selfsuperrl} pre-train the agent on randomly collected transitions and then perform  joint optimization of the policy and auxiliary losses. Importantly, the learning is done completely \textit{on-policy}: the policy loss is computed from rollouts while the auxiliary losses use samples from a small replay buffer. Yet, even with these precautions, the authors are unable to leverage reconstruction by VAE~\citep{kingma2013auto} and report its damaging affect on learning.

Similarly,~\citet{jaderberg2016unreal} propose to use unsupervised auxiliary tasks, both observation and reward based, and show improvements in Atari, again in an on-policy regime\footnote{\citet{jaderberg2016unreal} make use of a replay buffer that only stores the most recent 2K transitions, a small fraction of the 25M transitions experienced in training.}, which is much more stable for learning. Of all the auxiliary tasks considered by~\citet{jaderberg2016unreal}, reconstruction-based Pixel Control is the most effective. However, in maximizing changes in local patches, it imposes strong inductive biases that assume that dramatically changing pixel values and textures are correlated with good exploration and reward. Unfortunately, such highly task specific auxiliary is unlikely to scale to real world applications.

Generic pixel reconstruction is explored in \citet{higgins2017darla,nair2018imaginedgoal}, where the authors use a beta variational autoencoder ($\beta$-VAE)~\citep{kingma2013auto,higgins2017betavae} and attempt to perform joint representation learning, but find it hard to train, thus receding to the alternating training procedure~\citep{lange10deepaeinrl,finn2015deepspatialae}. 

 There has been more success in using model learning methods on images, such as \citet{hafner2018planet,lee2019slac}. These methods use a world model approach~\citep{ha2018worldmodels}, learning a representation space using a latent dynamics loss and pixel decoder loss to ground on the original observation space. These model-based reinforcement learning methods often show improved sample efficiency, but with the additional complexity of balancing various auxiliary losses, such as a dynamics loss, reward loss, and decoder loss in addition to the original policy and value optimizations. These proposed methods are correspondingly brittle to hyperparameter settings, and difficult to reproduce, as they balance multiple training objectives.

\section{Background}
\label{section:background}
\subsection{Markov Decision Process}
A fully observable Markov decision process (MDP) can be described as  $\gM=\langle\mathcal{S}, \mathcal{A}, P, R, \gamma\rangle$, where $\mathcal{S}$ is the state space, $\mathcal{A}$ is the action space, $P(\rvs_{t+1}|\rvs_t, \rva_t)$ is the transition probability distribution, $R(\rvs_t, \rva_t)$ is the reward function, and $\gamma$ is the discount factor~\citep{bellman1957mdp}. An agent starts in a initial state $\rvs_1$ sampled from a fixed distribution $p(\rvs_1)$, then at each timestep $t$ it takes an action $\rva_t \in \gA$ from a state $\rvs_t \in \gS$ and moves to a next state $\rvs_{t+1} \sim P(\cdot| \rvs_t, \rva_t)$. After each action the agent receives a reward $r_t = R(\rvs_{t}, \rva_t)$. We consider episodic environments with the length fixed to $T$. The goal of standard RL is to learn a policy $\pi(\rva_t|\rvs_t)$ that can maximize the agent's expected cumulative reward $\sum_{t=1}^T \E_{(\rvs_t, \rva_t) \sim \rho_\pi} [r_t]$, where $\rho_\pi$ is discounted state-action visitations of $\pi$, also known as occupancies. An important modification~\citep{ziebart2008maxent} auguments this objective with an entropy term $\mathcal{H}(\pi(\cdot|\rvs_t))$ to encourage exploration and robustness to noise. The resulting maximum entropy objective is then defined as
\begin{align}
    \pi^{*} &= \argmax_{\pi} \sum_{t=1}^T \E_{(\rvs_t, \rva_t) \sim \rho_\pi} [r_t + \alpha \gH(\pi(\cdot|\rvs_t))],
\end{align}
where $\alpha$ is temperature that balances between optimizing for the reward and for the stochasticity of the policy.
\subsection{Soft Actor-Critic}
Soft Actor-Critic (SAC)~\citep{haarnoja2018sac} is an \emph{off-policy} actor-critic method that uses the maximum entropy framework to derive soft policy iteration. At each iteration SAC performs soft policy evaluation and improvement steps. The policy evaluation step fits a parametric Q-function $Q(\rvs_t, \rva_t)$  using transitions sampled from the replay buffer $\gD$ by minimizing the soft Bellman residual
\begin{align}
    \label{eq:q_loss}
    J(Q) &= \E_{(\rvs_t, \rva_t, r_t, \rvs_{t+1}) \sim \gD} \bigg[ \big(Q(\rvs_t, \rva_t) - r_t - \gamma \Bar{V}(\rvs_{t+1})\big)^2  \bigg].
\end{align}
The target value function $\Bar{V}$ is approximated via a Monte-Carlo estimate of the following expectation
\begin{align}
    \Bar{V}(\rvs_{t}) &= \E_{\rva_{t} \sim \pi} \big[\Bar{Q}(\rvs_{t}, \rva_{t}) - \alpha  \log \pi(\rva_{t}|\rvs_{t}) \big],
\end{align}
where $\bar{Q}$ is the target Q-function parametrized by a weight vector obtained from an exponentially moving average of the Q-function weights to stabilize training. The  policy improvement step then attempts to project a parametric policy $\pi(\rva_t|\rvs_t)$  by minimizing KL divergence between the  policy and a Boltzmann distribution induced by the Q-function using the following objective
\begin{align}
    \label{eq:pi_loss}
    J(\pi) &= \E_{\rvs_t \sim \gD} \big[\KL(\pi(\cdot|\rvs_t) || \gQ(\rvs_t, \cdot)) \big],
\end{align}
where $\gQ(\rvs_t, \cdot) \propto \exp \{ \frac{1}{\alpha} Q(\rvs_t, \cdot) \}$.

\begin{figure*}[t!]

\centering
\includegraphics[width=0.16\textwidth]{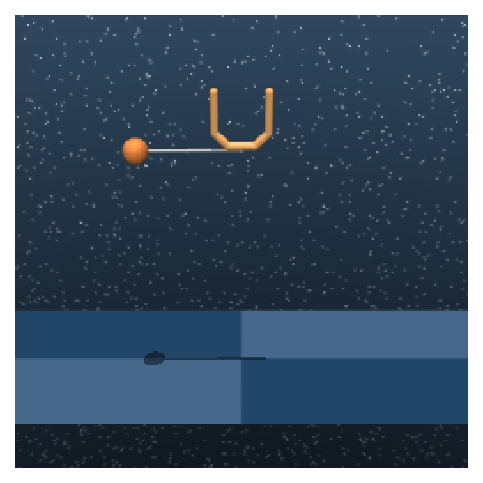}
\includegraphics[width=0.16\textwidth]{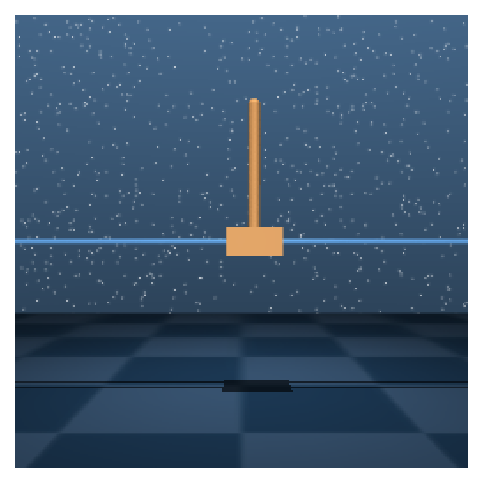}
\includegraphics[width=0.16\textwidth]{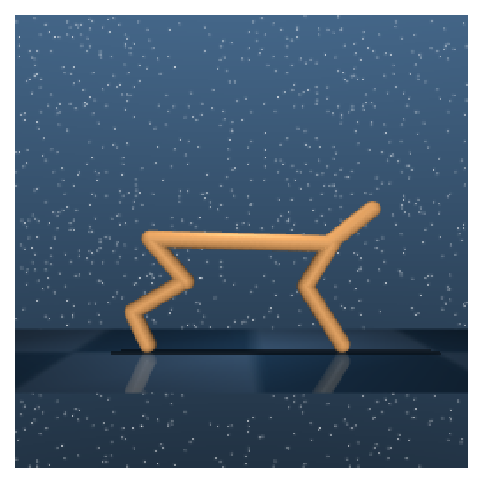}
\includegraphics[width=0.16\textwidth]{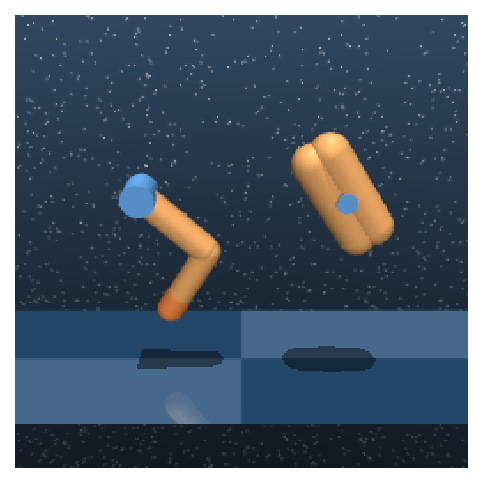}
\includegraphics[width=0.16\textwidth]{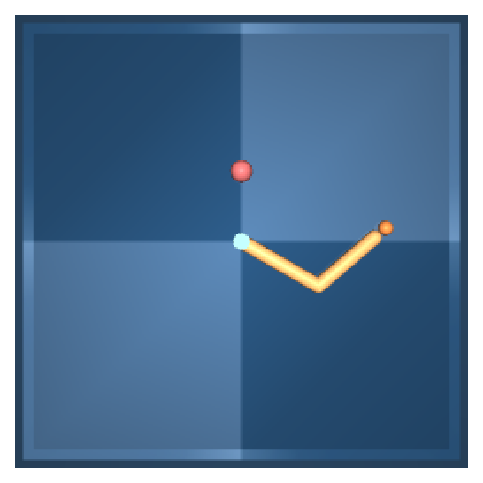}
\includegraphics[width=0.16\textwidth]{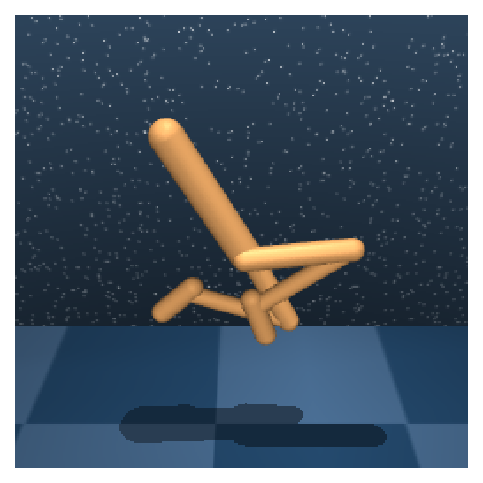}
    \vspace{-5pt}
\caption{Image-based continuous control tasks from the DeepMind Control Suite~\citep{tassa2018dmcontrol} used in our experiments. Each task offers an unique set of challenges, including complex dynamics, sparse rewards, hard exploration, and other traits (see~\cref{section:dm_control_suite}).}
\label{fig:dmc_domains_mini}
\end{figure*}

\subsection{Image-based Observations and Autoencoders}
Directly learning from raw images posses an additional problem of partial observability, which is formalized by a partially observable MDP (POMDP). In this setting, instead of getting a low-dimensional state $\rvs_t \in \gS$ at time $t$, the agent receives a high-dimensional observation $\rvo_t \in \gO$, which is a rendering of potentially incomplete view of the corresponding state $\rvs_t$ of the environment~\citep{kaelbling1998planning}. This complicates applying RL as the agent now needs to also learn a compact latent representation to infer the state. Fitting a high-capacity encoder using only a scarce reward signal is sample inefficient and prone to suboptimal convergence. Following prior work~\citep{lange10deepaeinrl,finn2015deepspatialae} we explore unsupervised pretraining via an image-based autoencoder (AE). In practice, the AE is represented as a convolutional encoder $g_{\phi}$ that maps an image observation $\rvo_t$ to a low-dimensional latent vector $\rvz_t$, and a deconvolutional decoder $f_{\theta}$ that reconstructs $\rvz_t$ back to the original image $\rvo_t$. Both the encoder and decoder are trained simultaneously by maximizing the expected log-likelihood
\begin{align}
    \label{eq:ae_loss}
    J(\mathrm{AE}) &= \E_{\rvo_t \sim \gD} \big[ \log p_{\theta} (\rvo_t | \rvz_t) \big],
\end{align}
where $\rvz_t = g_{\phi}(\rvo_t)$.
Or in the case of $\beta$-VAE~\citep{kingma2013auto,higgins2017betavae} we maximize the objective below
\begin{align}
    \label{eq:vae_loss}
    J(\mathrm{VAE}) &= \E_{\rvo_t \sim \gD} \big[\E_{\rvz_t \sim q_{\phi}(\rvz_t|\rvo_t)}[\log p_{\theta} (\rvo_t | \rvz_t)]\\\nonumber
    &- \beta \KL(q_{\phi}(\rvz_t|\rvo_t)||p(\rvz_t))  \big],  
\end{align}
where the variational distribution is parametrized as $q_{\phi}(\rvz_t|\rvo_t) = \gN(\rvz_t|\mu_{\phi}(\rvo_t), \sigma^2_{\phi}(\rvo_t))$.
The latent vector $\rvz_t$ is then used by an RL algorithm, such as SAC, instead of the unavailable true state $\rvs_t$. 

\section{Representation Learning with Image Reconstruction}
\label{section:dissection}

We start by noting a dramatic gap in an agent's performance when it learns from image-based observations rather than low-dimensional proprioceptive states.~\cref{table:mf_baseline} illustrates that in all cases \textbf{SAC:pixel} (an agent that learns from pixels) is significantly outperformed by \textbf{SAC:state} (an agent that learns from states). This result suggests that attaining a compact state representation is key in enabling efficient RL from images. Prior work has demonstrated that auxiliary supervision can improve representation learning, which is further confirmed in~\cref{table:mf_baseline} by superior performance of model-based methods, such as PlaNet~\citep{hafner2018planet} and SLAC~\citep{lee2019slac}, both of which make use of several auxiliary tasks to learn better representations.

While a wide range of auxiliary objectives could be added to aid effective representation learning, we focus our attention on the most general and widely applicable -- an image reconstruction loss. Furthermore, our goal is to develop a simple and robust algorithm that has the potential to be scaled up to real world applications (e.g. robotics). Correspondingly, we avoid task dependent auxiliary losses, such as Pixel Control from~\citet{jaderberg2016unreal}, or world-models~\citep{shelhamer16selfsuperrl,hafner2018planet,lee2019slac}. As noted by~\citet{gelada2019deepmdp} the latter can be brittle to train for reasons including: i) tension between reward and transition losses which requires careful tuning and  ii) difficulty in modeling complex dynamics (which we explore further in \cref{sec:noisy_observations}).

Following~\citet{nair2018imaginedgoal,hafner2018planet,lee2019slac}, which use reconstruction loss to learn the representation space and dynamics model with a variational autoencoder~\citep{kingma2013auto,higgins2017betavae}, we also employ a $\beta$-VAE  to learn representations, but in contrast to~\citet{hafner2018planet,lee2019slac} we only consider reconstructing the current frame, instead of reconstructing a temporal sequence of frames. Based on evidence from~\citet{lange10deepaeinrl,finn2015deepspatialae,nair2018imaginedgoal} we first try alternating between learning the policy and $\beta$-VAE, and in~\cref{section:dissection:pretraining} observe a positive correlation between the alternation frequency and the agent's performance. However, this approach does not fully close the performance gap, as the learned representation is not optimized for the task's objective. To address this shortcoming, we then attempt to additionally update the $\beta$-VAE encoder with the actor-critic gradients. Unfortunately, our investigation in~\cref{section:dissection:e2e} shows this approach to be ineffective due to severe instability in training, especially with larger $\beta$ values. Based on these results, in~\cref{section:dissection:stabilizing} we identify two reasons behind the instability, that originate from the stochastic nature of a $\beta$-VAE and the non-stationary gradient from the actor. We then propose two simple remedies and  in~\cref{section:dissection:sac-ae} introduce our method for an effective model-free off-policy RL from images.

\subsection{Experimental Setup}
Before carrying out our empirical study, we detail the experimental setup. A more comprehensive overview can be found in~\cref{section:hyperparams}. We evaluate all agents on six challenging control tasks (\cref{fig:dmc_domains_mini}). For brevity, on occasion, results for three tasks are shown  with the remainder presented in the appendix. An image observation is represented as a stack of three consecutive $84\times84$ RGB renderings~\citep{mnih2013dqn} to infer temporal statistics, such as velocity and acceleration. For simplicity, we keep the hyper parameters fixed across all the tasks, except for action repeat (see~\cref{section:hyperparams:training_setup}), which we set according to~\citet{hafner2018planet}
 for a fair comparison to the baselines. We evaluate an agent after every 10K training observations, by computing an average return over 10 episodes. For a reliable comparison we run 10 random seeds and report the mean and standard deviation of the evaluation reward. 


\subsection{Alternating Representation Learning with a $\beta$-VAE}

\label{section:dissection:pretraining}

\begin{figure}[h!]
    \centering
    \vspace{-10pt}
    \includegraphics[width=\linewidth]{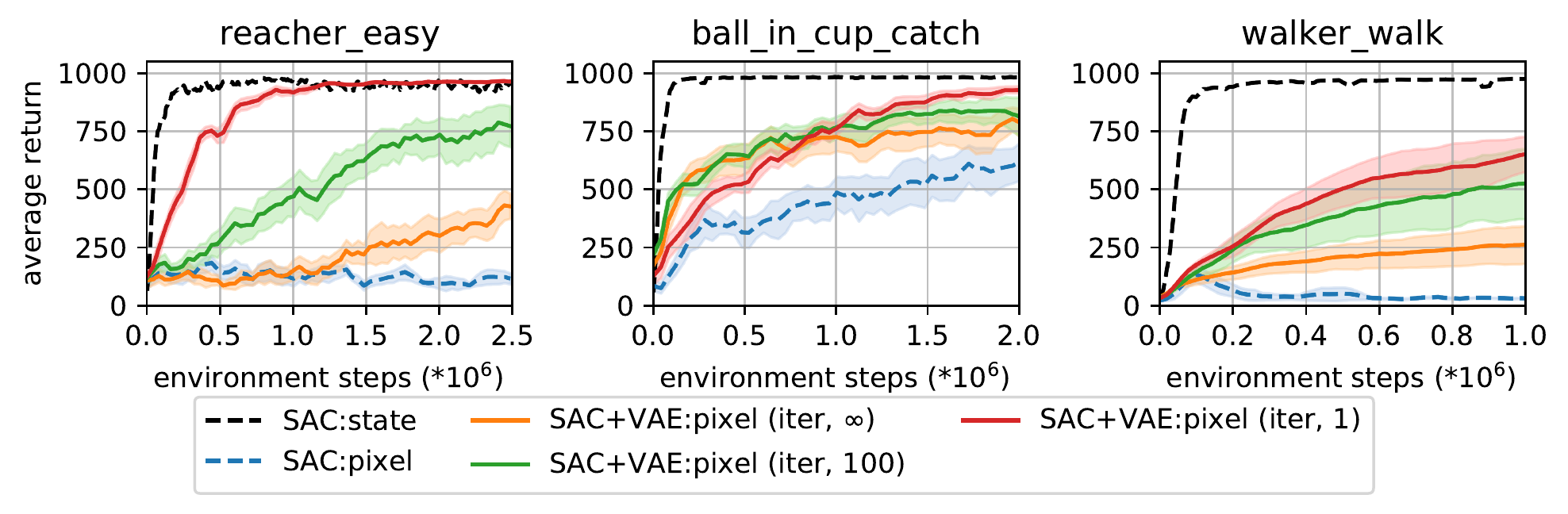}
    \vspace{-15pt}
    \caption{\small{Separate $\beta$-VAE and policy training with no shared gradients SAC+VAE:pixel (iter, $N$), with SAC:state shown as an upper bound. $N$ refers to frequency in environment steps at which the $\beta$-VAE updates after initial pretraining. More frequent updates are beneficial for learning better representations, but cannot fully address the gap in performance. Full results in~\cref{section:dissection:pretraining_full}. }}
    \label{fig:pretraining_small}
\end{figure}

We first set out to confirm the benefits of an alternating approach to representation learning in off-policy RL. We conduct an experiment where we initially pretrain the convolutional encoder $g_\phi$ and deconvolutional decoder $f_\theta$ of a $\beta$-VAE according to the loss $J(\mathrm{VAE})$ (\cref{eq:vae_loss}) on observations collected by a random policy. 
The actor and critic networks of SAC are then trained for $N$ steps using latent states $\rvz_t \sim g_\phi(\rvo_t)$ as inputs instead of image-based observations $\rvo_t$. We keep the encoder $g_\phi$ fixed during this period. The updated policy is then used to interact with the environment to gather new transitions that are consequently stored in the replay buffer. We continue iterating between the autoencoder and actor-critic updates until convergence. Note that the gradients are never shared between the $\beta$-VAE for learning the representation space, and the actor-critic.
In~\cref{fig:pretraining_small} we vary the frequency $N$ at which the representation space is updated, from $N=\infty$ where the representation is never updated after the initial pretraining period, to $N=1$ where the representation is updated after every policy update. We observe a positive correlation between this frequency and the agent's performance. Although the alternating scheme helps to improve the sample efficiency of the agent, it still falls short of reaching the upper bound performance of SAC:state. This is not surprising, as the learned representation space is never optimized for the task's objective.

\subsection{Joint Representation Learning with a $\beta$-VAE}
\label{section:dissection:e2e}
To further improve performance of the agent we seek to learn a latent representation that is well aligned with the underlying RL objective.   \citet{shelhamer16selfsuperrl} has demonstrated that joint policy and auxiliary objective optimization improves on the pretraining approach, as described in~\cref{section:dissection:pretraining}, but this has been only shown in the \emph{on-policy} regime. 

\begin{figure}[h!]
    \centering
    \includegraphics[width=\linewidth]{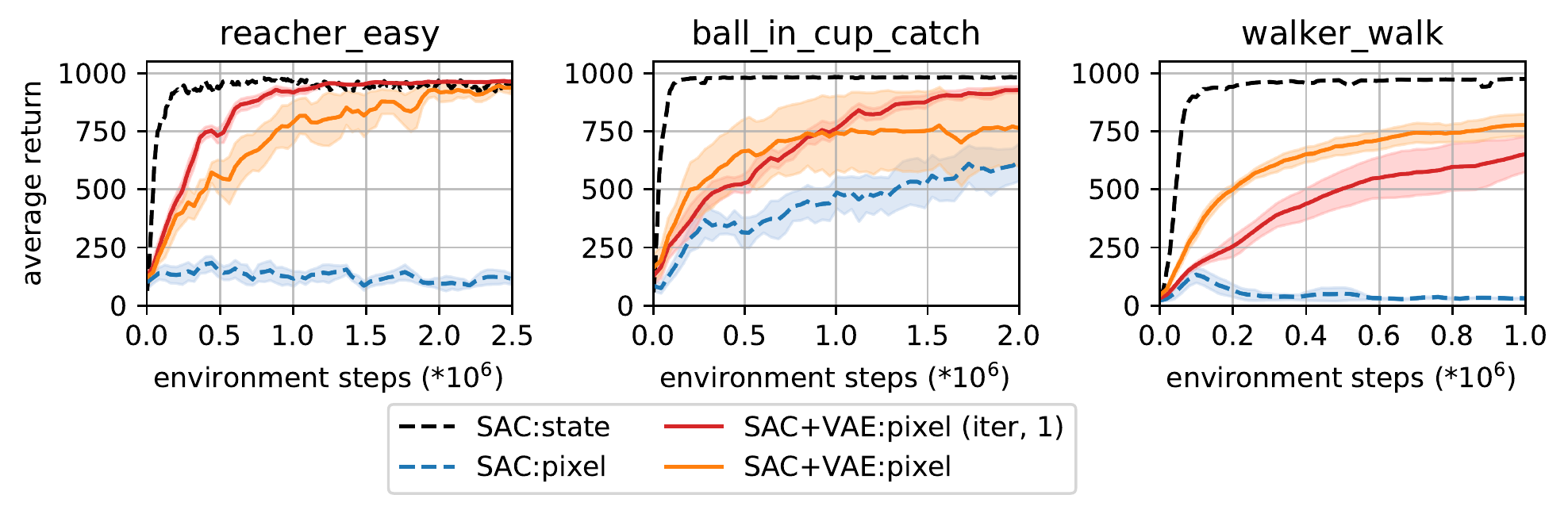}
    \vspace{-15pt}
    \caption{\small{An unsuccessful attempt to propagate gradients from the actor-critic down to the $\beta$-VAE encoder. SAC+VAE:pixel exhibits instability in training which leads to subpar performance comparing to the baseline SAC+VAE:pixel (iter, 1), which does not use the actor-critic gradients. Full results in~\cref{section:dissection:e2e_full}. }}
    \label{fig:ae_type_small}
\end{figure}

Thus we now attempt to verify the feasibility of joint representation learning with a $\beta$-VAE in the \emph{off-policy} setting. Specifically, we want to update the encoder network $g_\phi$ with the gradients coming through the latent state $\rvz_t$ from the actor $J(\pi)$ (\cref{eq:pi_loss}), critic $J(Q)$ (\cref{eq:q_loss}), and $\beta$-VAE $J(\mathrm{VAE})$ (\cref{eq:vae_loss}) losses. We thus take the best performing variant from the previous experiment (e.g. SAC+VAE:pixel (iter, 1)) and let the actor-critic's gradients update the encoder $g_\phi$. We tune for the best $\beta$ and name this agent \textbf{SAC+VAE:pixel}.
Results in~\cref{fig:ae_type_small} show that the joint representation learning with $\beta$-VAE in unstable in the \emph{off-policy} setting and performs worse than the baseline that does not utilize task dependent information (e.g. SAC+VAE:pixel (iter, 1)).

\subsection{Stabilizing Joint Representation Learning}
\label{section:dissection:stabilizing}

Following an unsuccessful attempt at joint representation learning with a $\beta$-VAE in off-policy RL, we investigate the root cause of the instability. 
\begin{figure}[h!]
    \centering
    \subfloat[Smaller values of $\beta$ reduce stochasticity of a $\beta$-VAE and lead to a better performance.  ]{\includegraphics[width=\linewidth]{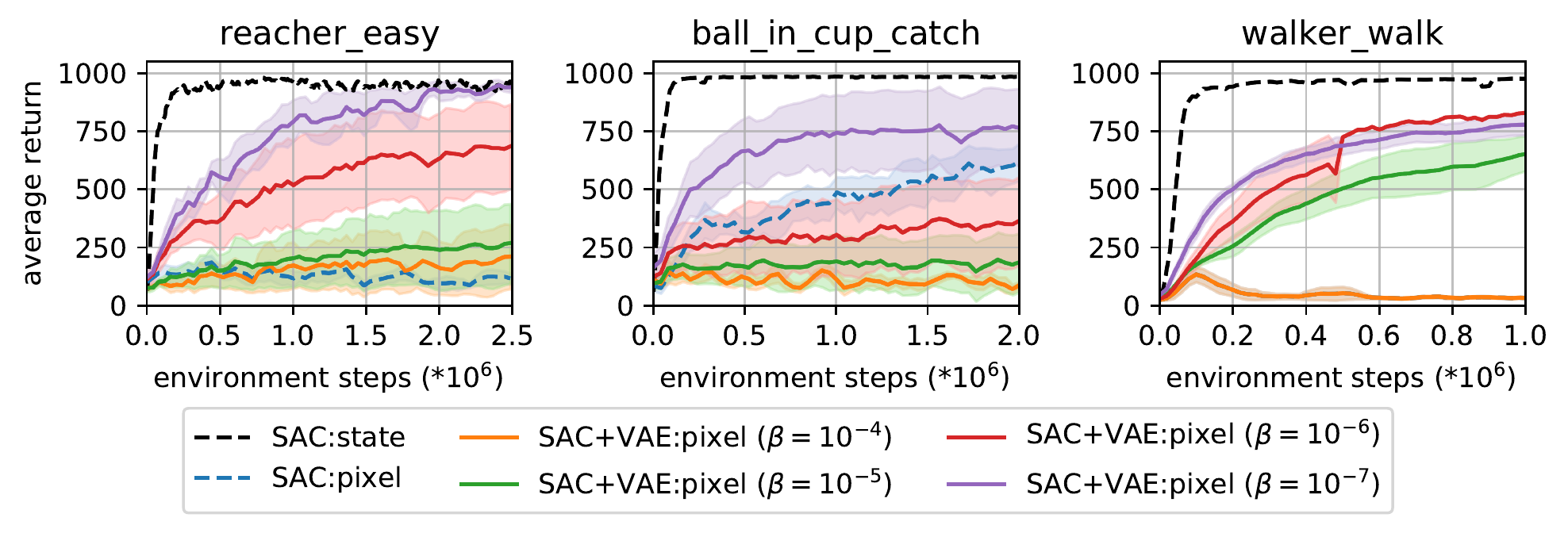} \label{fig:beta_vae}}\\
    \vspace{-10pt}
    \subfloat[Preventing the actor's gradients to update the convolutional encoder helps to improve performance even further.]{\includegraphics[width=\linewidth]{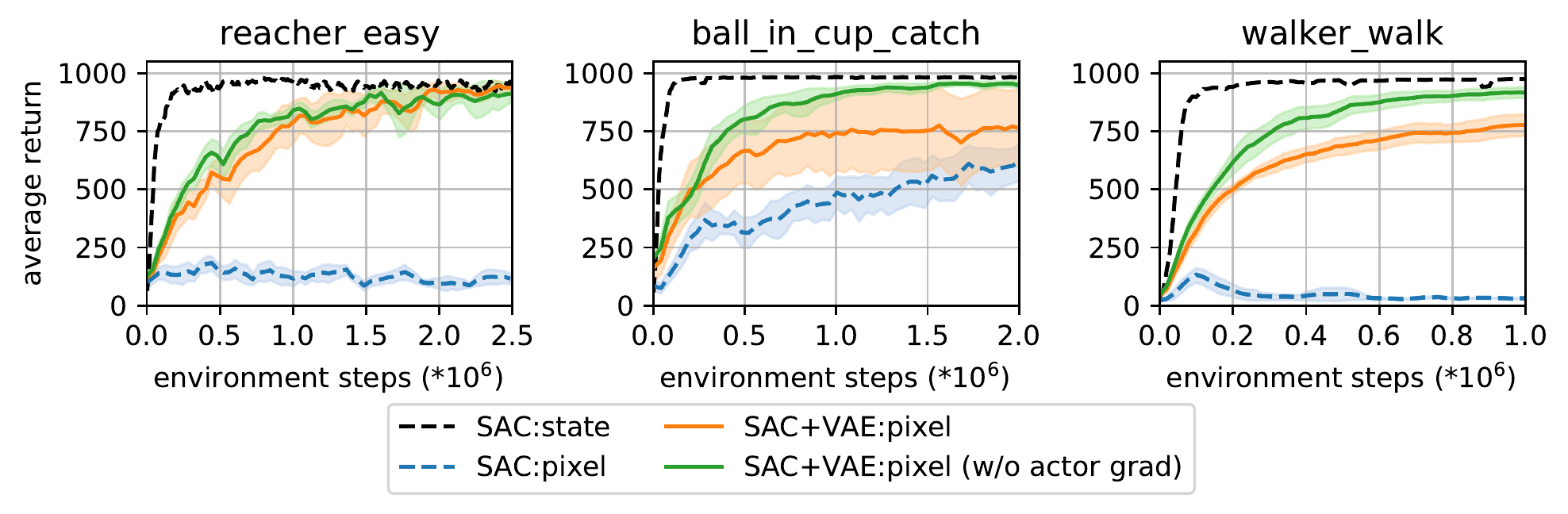}\label{fig:with_actor_grad}}
    \vspace{-10pt}
    \caption{\small{We identify two reasons for the subpar performance of joint representation learning. (a) The stochastic nature of a $\beta$-VAE, and (b) the non-stationary actor's gradients. Full results in~\cref{section:dissection:stabilizing_full}. }}
\end{figure}

\begin{figure*}[t!]
    \centering
    \includegraphics[width=0.7\linewidth]{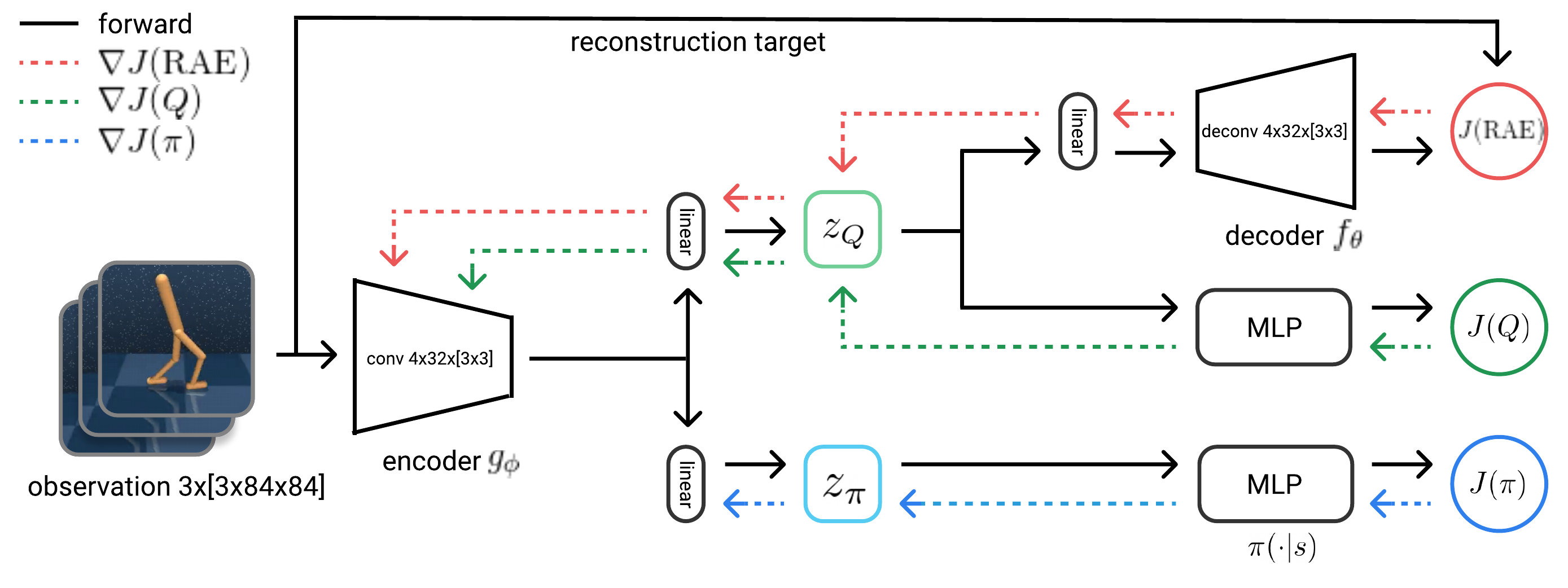}
    \vspace{-10pt}
    \caption{\small{Our algorithm (SAC+AE) auguments SAC with a regularized autoencoder to achieve stable training from images in the off-policy regime. The stability comes from switching to a deterministic encoder that is carefully updated with gradients from the reconstruction $J(\mathrm{RAE})$~(\cref{eq:rae_loss}) and soft Q-learning $J(Q)$~(\cref{eq:q_loss}) objectives.}}
    \label{fig:model}
\end{figure*}

We first observe that the stochastic nature of a $\beta$-VAE damages performance of the agent. The results from~\cref{fig:beta_vae} illustrate that smaller values of $\beta$ improve the training stability as well as the task performance. This motivates us to instead consider a completely deterministic autoencoder.

Furthermore, we observe that updating the convolutional encoder with the actor's gradients hurts the agent's performance. In~\cref{fig:with_actor_grad} we observe that blocking the actor's gradients from propagating to the encoder improves results considerably. This is because updating the encoder with the $J(\pi)$ loss (\cref{eq:pi_loss}) also changes the $Q$-function network inside the objective, due to the convolutional encoder being shared between the policy $\pi$ and $Q$-function. A similar phenomenon has been observed by~\citet{mnih2013dqn}, where the authors employ a static target $Q$-function to stabilize TD learning. It might appear that updating the encoder with only the critic's gradients would be insufficient to learn a task dependent representation space. However, the policy $\pi$ in SAC is a parametric projection of a Boltzmann distribution induced by the $Q$-function, see~\cref{eq:pi_loss}.
Thus, the $Q$-function contains all relevant information about the task and allows the encoder to learn task dependent representations from the critic's gradient alone.

\subsection{Our Approach SAC+AE: Joint Off-Policy Representation Learning}
\label{section:dissection:sac-ae}

We now introduce our approach \textbf{SAC+AE} -- a stable off-policy RL algorithm from images, derived from the above findings. We first replace the $\beta$-VAE with a deterministic autoencoder. 
To preserve the regularization affects of a $\beta$-VAE we adopt the RAE approach of~\citet{ghosh2019rae}, which 
imposes a $L_2$ penalty on the learned representation $\rvz_t$ and weight-decay on the decoder parameters
\begin{align}
    J(\mathrm{RAE})&= \E_{\rvo_t \sim \gD} \big[ \log p_{\theta}(\rvo_t|\rvz_t) + \lambda_{\rvz} ||\rvz_t||^2 + \lambda_{\theta} ||\theta||^2 \big],
    \label{eq:rae_loss}
\end{align}
where $\rvz_t=g_{\phi}(\rvo_t)$, and $\lambda_{\rvz}$, $\lambda_{\theta}$ are hyper parameters.

We also prevent the actor's gradients from updating the convolutional encoder, as suggested in~\cref{section:dissection:stabilizing}. Unfortunately, this slows down signal propogation to the encoder, and thus we find it important to update the convolutional weights of the target $Q$-function faster than the rest of the network's parameters. We thus employ different rates $\tau_Q$ and $\tau_\mathrm{enc}$ (with $\tau_\mathrm{enc} > \tau_Q$) to compute Polyak averaging over the corresponding parameters of the target $Q$-function. Our approach is summarized in~\cref{fig:model}.

\section{Evaluation of SAC+AE}

\begin{figure*}[t!]
    \centering
    \subfloat[Our method demonstrates significantly improved performance over the baseline  SAC:pixel. Moreover, it matches the state-of-the-art performance of world-model based algorithms, such as PlaNet and SLAC, as well as a model-free algorithm D4PG, that learns directly from raw images. Our algorithm is extremely stable, robust, and straightforward to implement.  ]{\includegraphics[width=\linewidth]{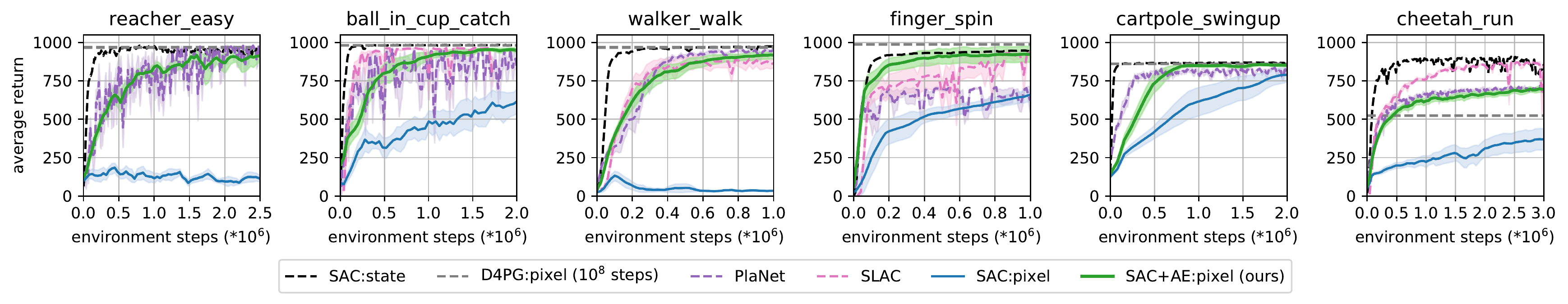}\label{fig:benchmark_planet}}\\
    \vspace{-10pt}
    \subfloat[Methods that rely on forward modeling, such as PlaNet and SLAC, suffer severely from the background noise, while our approach is resistant to the distractors. Examples of background distractors are show in~\cref{fig:distractors_small}.]{\includegraphics[width=\linewidth]{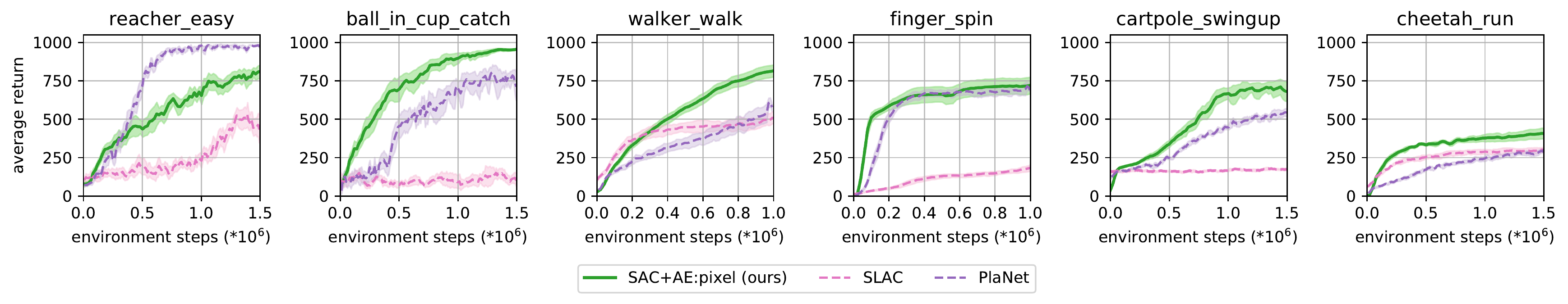}\label{fig:distractors_results}}
    \vspace{-5pt}
    \caption{\small{Two main results of our work. In (a) we demonstrate that our simple method matches the state-of-the-art performance on DMC tasks. In (b) we outperform the baselines on more complicated tasks where the observations are altered with noise.  }}
    
\end{figure*}

In this section we evaluate our approach, \textbf{SAC+AE}, on various benchmark tasks and compare against state-of-the-art methods, both model-free and model-based.  We then highlight the benefits of our model-free approach over those model-based methods in modified environments with distractors, as an approximation of real world noise. Finally, we test generalization to unseen tasks and dissect the representation power of the encoder.

\subsection{Learning Control from Pixels} 
We evaluate our method on six challenging image-based continuous control tasks (see~\cref{fig:dmc_domains_mini}) from DMC~\citep{tassa2018dmcontrol} and compare 
against several state-of-the-art model-free and model-based RL algorithms for learning from pixels: \textbf{D4PG}~\citep{barth-maron2018d4pg}, an off-policy actor-critic algorithm; \textbf{PlaNet}~\citep{hafner2018planet}, a model-based method that learns a dynamics model with deterministic and stochastic latent variables and employs cross-entropy planning for control; and \textbf{SLAC}~\citep{lee2019slac}, which combines a purely stochastic latent model together with an model-free soft actor-critic. In addition, we compare against SAC:state that learns from low-dimensional proprioceptive state, as an upper bound on performance. 
Results in~\cref{fig:benchmark_planet} illustrate that \textbf{SAC+AE:pixel} matches  the state-of-the-art model-based methods such as PlaNet and SLAC, despite being extremely simple and straightforward to implement.

\subsection{Performance on Noisy Observations}
\label{sec:noisy_observations}
Performing accurate forward-modeling predictions based off of noisy observations is challenging and requires learning a high fidelity model that encapsulates strong inductive biases~\citep{watters2017vin}. The current state-of-the-art world-model based approaches~\citep{hafner2018planet,lee2019slac} solely rely on a general purpose recurrent state-space model parametrized with a $\beta$-VAE, and thus are highly vulnerable to the observational noise. In contrast, the representations learned with just reconstruction loss are better suited to handle the background noise.

\begin{figure}[t!]
    \centering
    \includegraphics[width=0.15\textwidth]{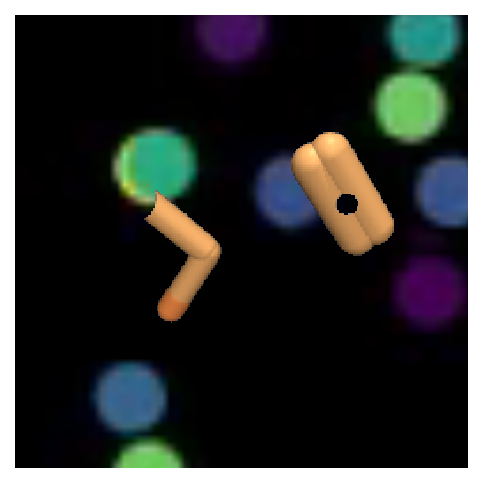}
    \includegraphics[width=0.15\textwidth]{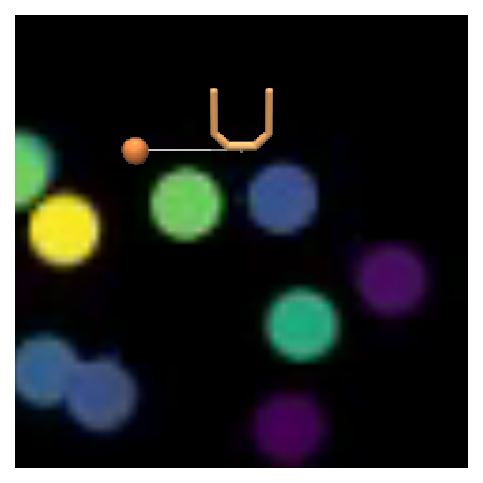}
    \includegraphics[width=0.15\textwidth]{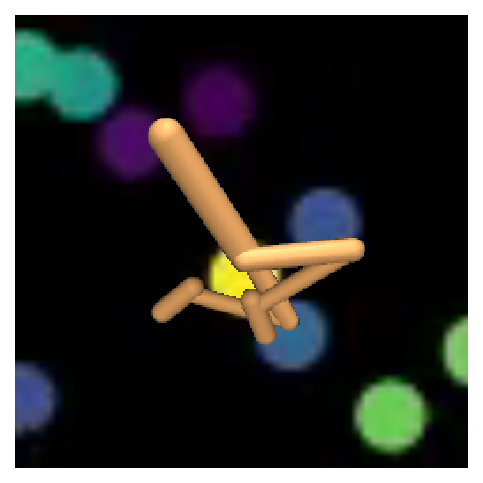}
   \vspace{-10pt}
    \caption{Backgrounds altered with randomly moving distractors.}
    \vspace{-15pt}
    \label{fig:distractors_small}
\end{figure}

\begin{figure}[t!]
    \centering
    \vspace{-10pt}
    \includegraphics[width=\linewidth]{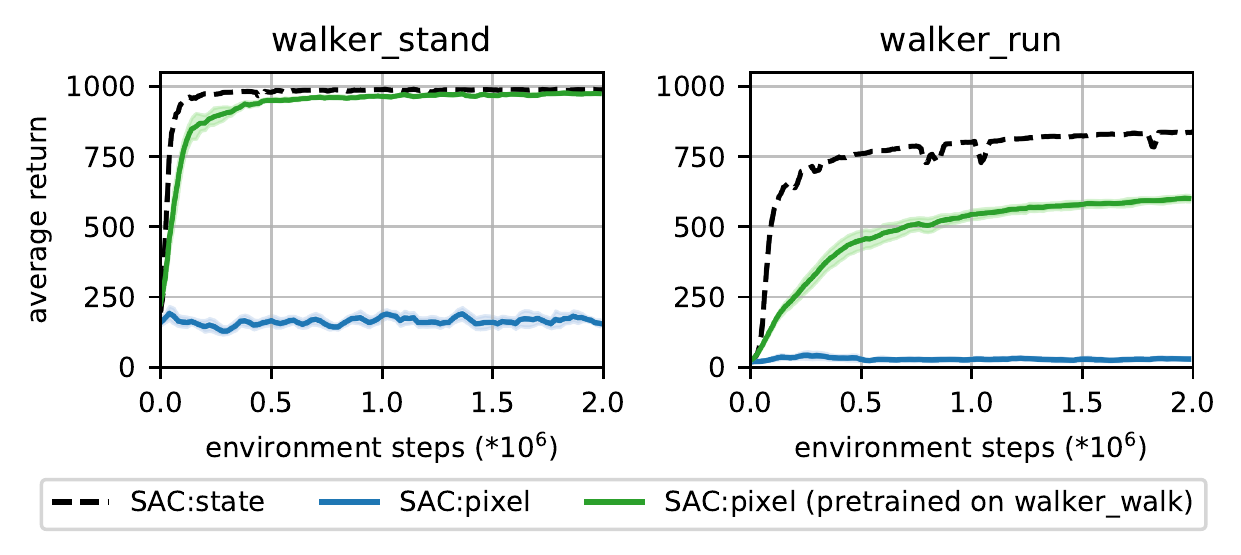}
    \vspace{-15pt}
    \caption{\small{An encoder pretrained with our method (SAC+AE:pixel) on \texttt{walker\_walk} is able to generalize to unseen \texttt{walker\_stand} and \texttt{walker\_run} tasks. All three tasks share similar image observations, but have quite different reward structure. SAC with a pretrained on \texttt{walker\_walk} encoder significantly outperforms the baseline. }}
    \label{fig:multitasking}
\end{figure}

To confirm this, we evaluate several agents on tasks where we add simple distractors in the background, consisting of colored balls bouncing off each other and the frame (\cref{fig:distractors_small}). We use image processing to filter away the static background and replace it with this dynamic noise, as proposed in \citet{zhang2018natrl}. We aim to emulate a common setup in a robotics lab, where various unrelated objects can affect robot's observations. In~\cref{fig:distractors_results} we see that methods that rely on forward modeling perform drastically worse than our approach, showing that our method is more robust to background noise.

\begin{figure*}[t!]
    \centering
    \includegraphics[width=0.7\linewidth]{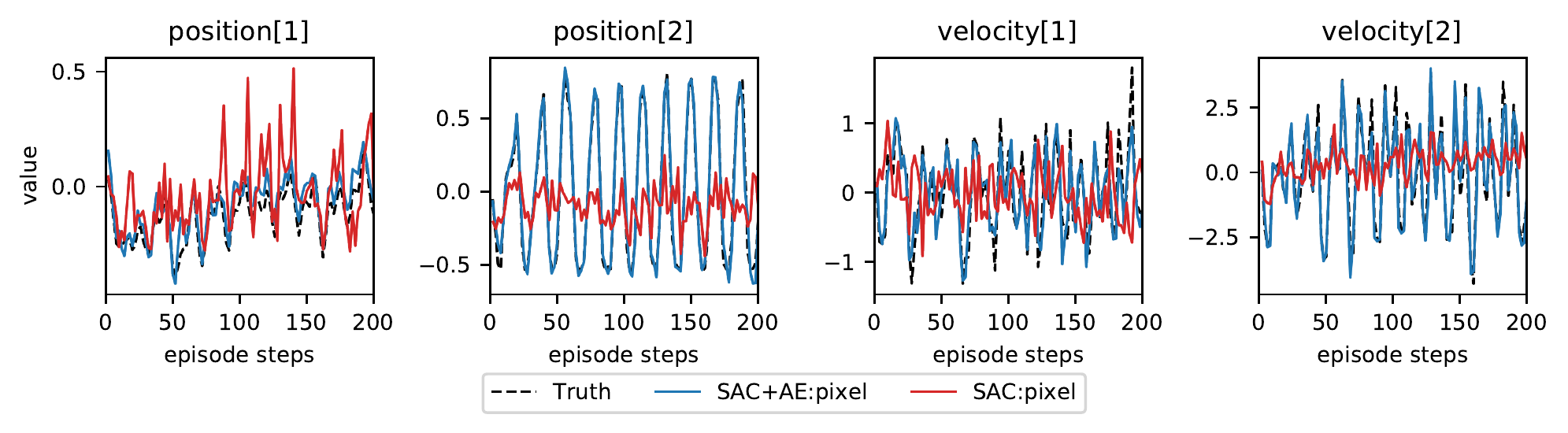}
    \vspace{-10pt}
    \caption{\small{Linear projections of latent representation spaces learned by our method (SAC+AE:pixel) and the baseline (SAC:pixel) onto proprioceptive states. We compare ground truth value of each proprioceptive coordinate against their reconstructions for \texttt{cheetah\_run}, and conclude that our method successfully encodes proprioceptive state information.  For visual clarity we only plot $2$ position (out of $8$) and $2$ velocity (out of $9$) coordinates. Full results in~\cref{section:ablation:repr_power_full}. }}
    \vspace{-10pt}
    \label{fig:encoder_proprio_rec_small}
\end{figure*}

\subsection{Generalization to Unseen Tasks}
Next, we show that the latent representation space learned by our method is able to generalize to different tasks without additional fine-tuning. We take three tasks \texttt{walker\_stand}, \texttt{walker\_walk}, and \texttt{walker\_run} from DMC, which share the same observational appearance, but all have different reward functions. We train SAC+AE:pixel on the \texttt{walker\_walk} task until convergence and fix the encoder. Consequently, we train two SAC:pixel agents on \texttt{walker\_stand} and \texttt{walker\_run}. The encoder of the first agent is initialized with weights from the pretrained on \texttt{walker\_walk} encoder, while the encoder of the second agent is not. Neither of the agents uses the reconstruction signal, and only backpropogate the critic's gradients. Results in~\cref{fig:multitasking} illustrate that our method learns latent representations that can readily generalize to unseen tasks and achieve much better performance than SAC:pixel trained from scratch.

\subsection{Representation Power of the Encoder}
\label{section:ablation:repr_power}

Finally, we want to determine if our method is able to extract sufficient information from raw images to recover the corresponding proprioceptive states. 
We thus train SAC+AE:pixel and SAC:pixel until convergence on \texttt{cheetah\_run} and then fix their encoders. We then learn two linear projections to map the encoders' latent embedding of image observations into the corresponding proprioceptive states.   Finally, we compare ground truth proprioceptive states against their reconstructions. We emphasize that the image encoder attributes for over $90\%$ of the agent's parameters, thus we believe that the encoder's latent output $\rvz_t$ captures a significant amount of information about the corresponding internal state in both cases, even though SAC:pixel does not require this explicitly. Results in~\cref{fig:encoder_proprio_rec_small} confirm that the internals of the task are easily extracted from the encoder grounded on pixel observations, whereas they are much more difficult to construct from the representation learned by SAC:pixel.


\section{Discussion}
For RL agents to be effective in the  real world, where vision is one of the richest sensing modalities, we need sample efficient, robust algorithms that work from pixel observations. We pinpoint two strategies to obtain sample efficiency -- i) use off-policy methods and ii) use self-supervised auxiliary losses. For methods to be robust, we want auxiliary losses that do not rely on task-specific inductive biases, so we focus on a simple reconstruction loss.
In this work, we provide a thorough study into combining reconstruction loss with off-policy methods for improved sample efficiency in rich observation settings. 
Our analysis yields two key findings. The first is that deterministic AE models outperform $\beta$-VAEs~\citep{higgins2017betavae}, due to additional instabilities such as bootstrapping, off-policy data, and joint training with auxiliary losses. The second is that propagating the actor's gradients through the convolutional encoder hurts performance.

Based on these results, we also recommend an effective off-policy, model-free RL algorithm for pixel observations with only reconstruction loss as an auxiliary task. It is competitive with state-of-the-art model-based methods on traditional benchmarks, but much simpler, robust, and does not require learning a dynamics model (\cref{fig:benchmark_planet}). We show through ablations the superiority of joint learning over previous methods that use an alternating training procedure with separated gradients, the necessity of a pixel reconstruction loss over reconstruction to lower-dimensional ``correct" representations, and demonstrations of the representation power and generalization ability of our learned representation. We additionally construct settings with distractors approximating real world noise which show how learning a world-model as an auxiliary loss can be harmful (\cref{fig:distractors_results}), and in which our method, SAC+AE, exhibits state-of-the-art performance.

In the Appendix we provide results across all experiments on the full suite of 6 tasks chosen from DMC (\cref{section:dm_control_suite}), and the full set of hyperparameters used in~\cref{section:hyperparams}. There are also additional experiments  autoencoder capacity (\cref{section:autoencoder_capacity}), a look at optimality of the learned latent representation (\cref{section:optimality}) and importance of action repeat (\cref{section:action_repeat}). Finally, we opensource our codebase for the community to spur future research in image-based RL. 

\newpage 

\bibliography{main}
\bibliographystyle{icml2020}

\includeappendixtrue 

\ifincludeappendix

\newpage
\onecolumn
\appendix
\section*{Appendix}
\section{The DeepMind Control Suite}
\label{section:dm_control_suite}
We evaluate the algorithms in the paper on the DeepMind control suite (DMC)~\citep{tassa2018dmcontrol} -- a collection of continuous control tasks that offers an excellent testbed for reinforcement learning agents.
The software emphasizes the importance of having a standardised set of benchmarks with a unified reward structure in order to measure made progress reliably. 

Specifically, we consider six domains (see~\cref{fig:dmc_domains}) that result in twelve different control tasks. Each task (\cref{table:dm_control_tasks}) poses a particular set of challenges to a learning algorithm. The \texttt{ball\_in\_cup\_catch} task only provides the agent with a sparse reward when the ball is caught; the \texttt{cheetah\_run} task offers high dimensional internal state and action spaces; the \texttt{reacher\_hard} task requires the agent to explore the environment. We refer the reader to the original paper to find more information about the benchmarks.

\begin{table}[!b]
\centering
\resizebox{0.8\textwidth}{!}{
\begin{tabular}{|l|cc|c|c|}
\hline
Task name        & \multicolumn{2}{c|
}{$\mathrm{dim}(\gO)$} & $\mathrm{dim}(\gA)$ & Reward type     \\
& Proprioceptive & Image-based & & \\
\hline
\texttt{ball\_in\_cup\_catch} & $8$  & $3\times84\times84$& $2$ & sparse \\
\texttt{cartpole\_\{balance,swingup\}} & $5$ &$3\times84\times84$ &$1$  & dense \\
\texttt{cheetah\_run} & $17$ &$3\times84\times84$ & $6$ & dense \\
\texttt{finger\_\{spin,turn\_easy,turn\_hard\}} & $12$ &$3\times84\times84$ & $2$ & dense/sparse \\
\texttt{reacher\_\{easy,hard\}} & $7$ & $3\times84\times84$& $2$& sparse \\
\texttt{walker\_\{stand,walk,run\}} & $24$&$3\times84\times84$ & $6$ & dense \\

\hline
\end{tabular}}
\caption{\label{table:dm_control_tasks} Specifications of observation space $\gO$ (proprioceptive and image-based), action space $\gA$, and the reward type for each task.}
\end{table}

\begin{figure*}[!b]

\centering
\subfloat[\texttt{ball\_in\_cup}]{
\includegraphics[width=0.25\textwidth]{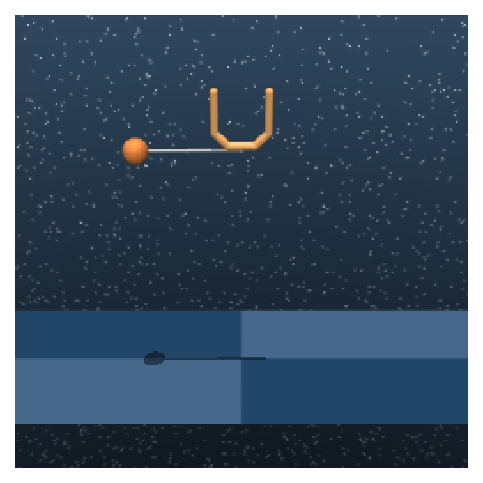}}
\subfloat[\texttt{cartpole}]{
\includegraphics[width=0.25\textwidth]{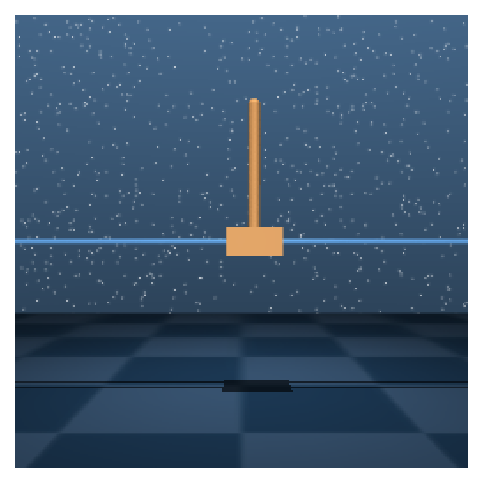}}
\subfloat[\texttt{cheetah}]{
\includegraphics[width=0.25\textwidth]{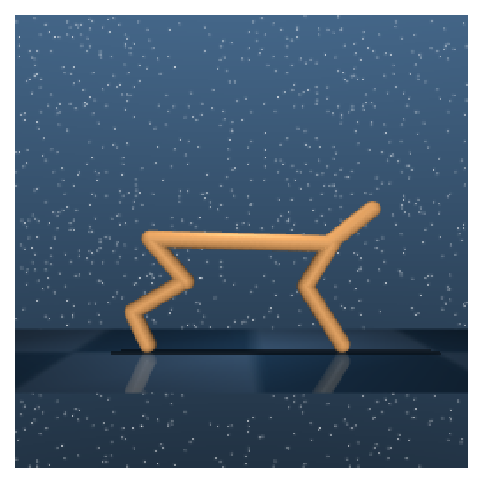}}

\subfloat[\texttt{finger}]{
\includegraphics[width=0.25\textwidth]{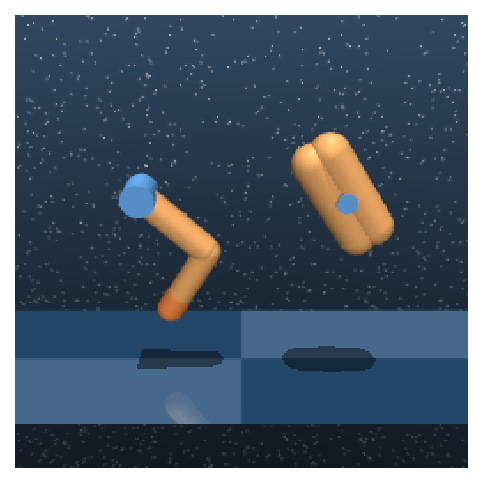}}
\subfloat[\texttt{reacher}]{
\includegraphics[width=0.25\textwidth]{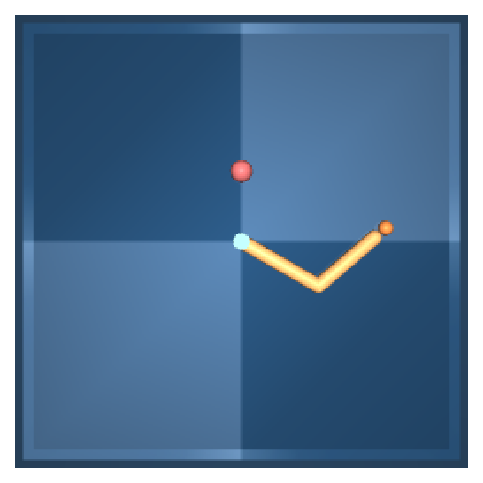}}
\subfloat[\texttt{walker}]{
\includegraphics[width=0.25\textwidth]{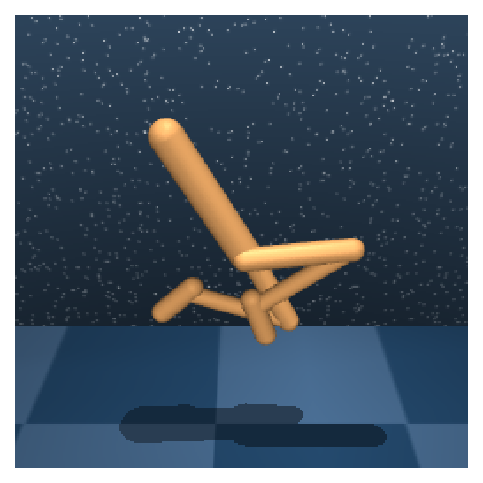}}

\caption{Our testbed consists of six domains spanning the total of twelve challenging continuous control tasks: \texttt{finger\_\{spin,turn\_easy,turn\_hard\}}, \texttt{cartpole\_\{balance,swingup\}}, \texttt{cheetah\_run}, \texttt{walker\_\{stand,walk,run\}}, \texttt{reacher\_\{easy,hard\}}, and \texttt{ball\_in\_cup\_catch}.}
\label{fig:dmc_domains}
\end{figure*}

\newpage

\section{Hyper Parameters and Setup}
\label{section:hyperparams}

Our PyTorch SAC~\cite{haarnoja2018sac} implementation is based off of~\cite{yarats2020pytorch_sac}.

\subsection{Actor and Critic Networks}
We employ double Q-learning~\citep{hasselt2015doubledqn} for the critic, where each Q-function is parametrized as a 3-layer MLP with \texttt{ReLU} activations after each layer except of the last. The actor is also a 3-layer MLP with \texttt{ReLU}s that outputs mean and covariance for the diagonal Gaussian that represents the policy. The hidden dimension is set to $1024$ for both the critic and actor.

\subsection{Encoder and Decoder Networks}
We employ an almost identical encoder architecture as in~\citet{tassa2018dmcontrol}, with two minor differences. Firstly, we add two more convolutional layers to the convnet trunk. Secondly, we use \texttt{ReLU} activations after each conv layer, instead of \texttt{ELU}. We employ kernels of size $3 \times 3$ with $32$ channels  for all the conv layers and set stride to $1$ everywhere, except of the first conv layer, which has stride $2$. We then take the output of the convnet and feed it into a single fully-connected layer normalized by \texttt{LayerNorm}~\citep{ba2016layernorm}. Finally, we add \texttt{tanh} nonlinearity to the $50$ dimensional output of the fully-connected layer.

The actor and critic networks both have separate encoders, although we share the weights of the conv layers between them. Furthermore, only the critic optimizer is allowed to update these weights (e.g. we truncate the gradients from the actor before they propagate to the shared conv layers).

The decoder consists of one fully-connected layer that is then followed by four deconv layers. We use \texttt{ReLU} activations after each layer, except the final deconv layer that produces pixels representation. Each deconv layer has kernels of size $3 \times 3$ with $32$ channels and stride $1$, except of the last layer, where stride is $2$.

We then combine the critic's encoder together with the decoder specified above into an autoencoder. Note, because we share conv weights  between the critic's and actor's encoders, the conv layers of the actor's encoder will be also affected by reconstruction signal from the autoencoder.

\subsection{Training and Evaluation Setup}
\label{section:hyperparams:training_setup}


We first collect $1000$ seed observations using a random policy. We then collect training observations by sampling actions from the current policy. We perform one training update every time we receive a new observation. In cases where we use action repeat, the number of training observations is only a fraction of the environment steps (e.g. a $1000$ steps episode at action repeat $4$ will only results into $250$ training observations). The action repeat used for each environment is specified in \cref{table:action_repeat}, following those used by PlaNet and SLAC.

We evaluate our agent after every $10000$ environment steps by computing an average episode return over $10$ evaluation episodes. 
Instead of sampling from the Gaussian policy we take its mean during evaluation.

We preserve this setup throughout all the experiments in the paper.
\begin{table}[ht!]
\centering
\begin{tabular}{|l|c|}
\hline
Task name        & Action repeat \\
\hline
\texttt{cartpole\_swingup} &  8 \\
\texttt{reacher\_easy} &  4 \\
\texttt{cheetah\_run} & 4 \\
\texttt{finger\_spin} & 2 \\
\texttt{ball\_in\_cup\_catch} & 4 \\
\texttt{walker\_walk} & 2 \\
\hline
\end{tabular}\\
\caption{\label{table:action_repeat} Action repeat parameter used per task, following PlaNet and SLAC.}
\end{table}

\subsection{Weights Initialization}
We initialize the weight matrix of fully-connected layers with the orthogonal initialization~\citep{saxe2013ortho} and set the bias to be zero. For convolutional and deconvolutional layers we use delta-orthogonal initialization~\citep{xiao2018deltainit}.

\subsection{Regularization}
\label{section:hyperparams:regularization}
We regularize the autoencoder network using the scheme proposed in~\citet{ghosh2019rae}. In particular, we extend the standard reconstruction loss for a deterministic autoencoder with  a $L_2$ penalty on the learned representation $\rvz$ and add weight decay on the decoder parameters $\theta$:
\begin{align}
    J(\mathrm{RAE})&= \E_{\rvo_t \sim \gD} \big[ \log p_{\theta}(\rvo_t|\rvz_t) + \lambda_{\rvz} ||\rvz_t||^2 + \lambda_{\theta} ||\theta||^2 \big] \quad\text{where}\quad \rvz_t=g_{\phi}(\rvo_t).
\end{align}
We set $\lambda_z=10^{-6}$ and $\lambda_\theta = 10^{-7}$.

\subsection{Pixels Preprocessing}
We construct an observational input as an $3$-stack of consecutive frames~\citep{mnih2013dqn}, where each frame is a RGB rendering of size $ 84 \times 84$ from the $0$th camera. We then divide each pixel by $255$ to scale it down to $[0, 1)$ range. For reconstruction targets we instead preprocess images by reducing bit depth to 5 bits as in \citet{kingma2018glow}.

\subsection{Other Hyper Parameters}
We also provide a comprehensive overview of all the remaining hyper parameters in~\cref{table:hyper_params}.

\begin{table}[hb!]
\centering
\begin{tabular}{|l|c|}
\hline
Parameter name        & Value \\
\hline
Replay buffer capacity & $1000000$ \\
Batch size & $128$ \\
Discount $\gamma$ & $0.99$ \\
Optimizer & Adam \\
Critic learning rate & $10^{-3}$ \\
Critic target update frequency & $2$ \\
Critic Q-function soft-update rate $\tau_{\textrm{Q}}$ & 0.01 \\
Critic encoder soft-update rate $\tau_{\textrm{enc}}$ & 0.05 \\
Actor learning rate & $10^{-3}$ \\
Actor update frequency & $2$ \\
Actor log stddev bounds & $[-10, 2]$ \\
Autoencoder learning rate & $10^{-3}$ \\
Temperature learning rate & $10^{-4}$ \\
Temperature Adam's $\beta_1$ & $0.5$ \\
Init temperature & $0.1$ \\

\hline
\end{tabular}\\
\caption{\label{table:hyper_params} A complete overview of used hyper parameters.}
\end{table}

\newpage

\section{Alternating Representation Learning with a $\beta$-VAE}
\label{section:dissection:pretraining_full}
Iterative pretraining suggested in~\citet{lange10deepaeinrl,finn2015deepspatialae} allows for faster representation learning, which consequently boosts the final performance, yet it is not sufficient enough to fully close the gap and additional modifications, such as joint training, are needed.~\cref{fig:pretraining} provides additional results for the experiment described in~\cref{section:dissection:pretraining}.

\begin{figure}[hb!]
    \centering
    \includegraphics[width=0.9\linewidth]{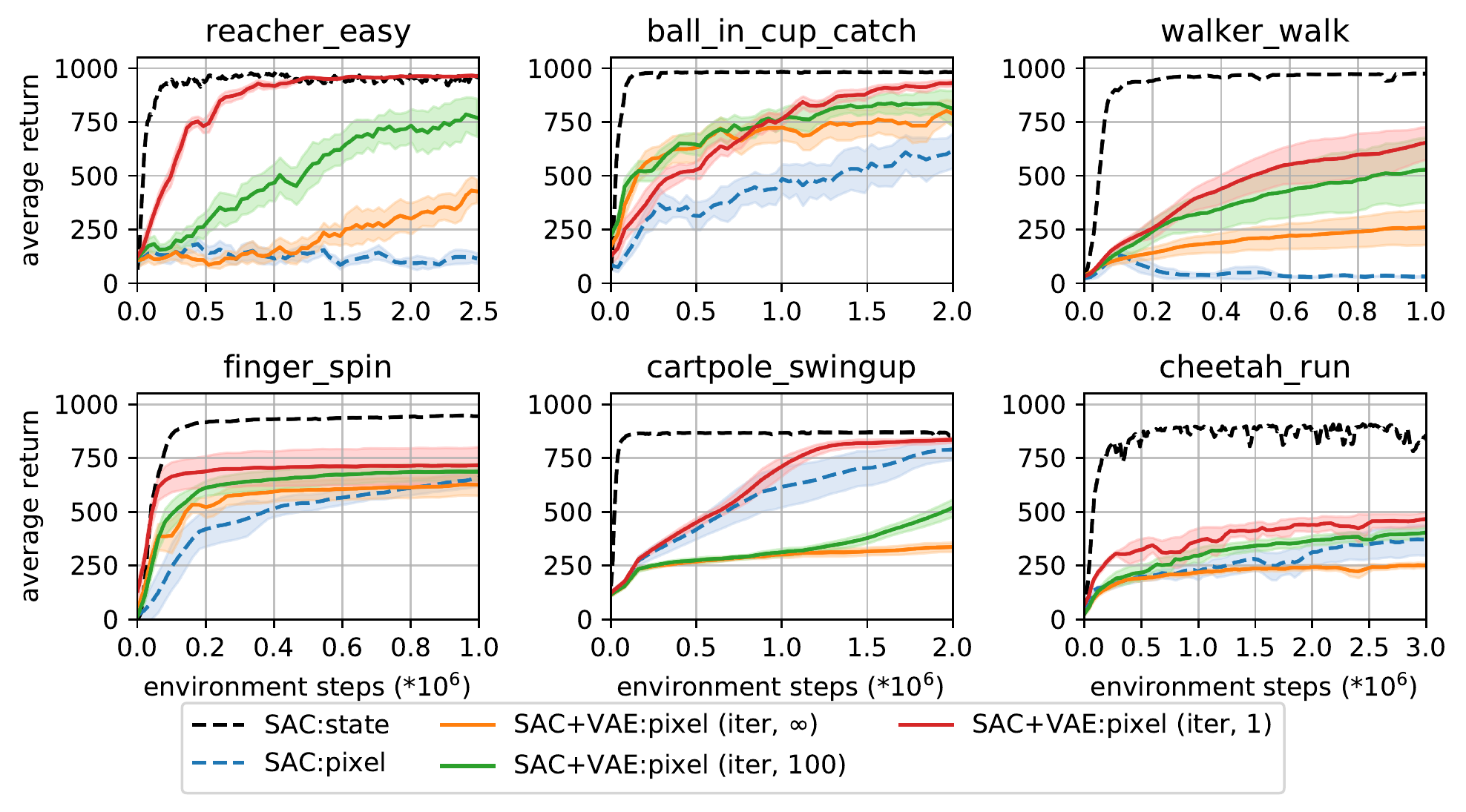}
      \vspace{-5pt}
    \caption{Separate $\beta$-VAE and policy training with no shared gradients SAC+VAE:pixel (iter, $N$), with SAC:state shown as an upper bound. $N$ refers to frequency in environment steps at which the $\beta$-VAE updates after initial pretraining. More frequent updates are beneficial for learning better representations, but cannot fully address the gap in performance.  }
    \label{fig:pretraining}
    
\end{figure}

\newpage

\section{Joint Representation Learning with a $\beta$-VAE}
\label{section:dissection:e2e_full}
Additional results to the experiments from~\cref{section:dissection:e2e} are in~\cref{fig:beta_vae} and~\cref{fig:ae_type}.

\begin{figure}[ht!]
    \centering
    \includegraphics[width=0.9\linewidth]{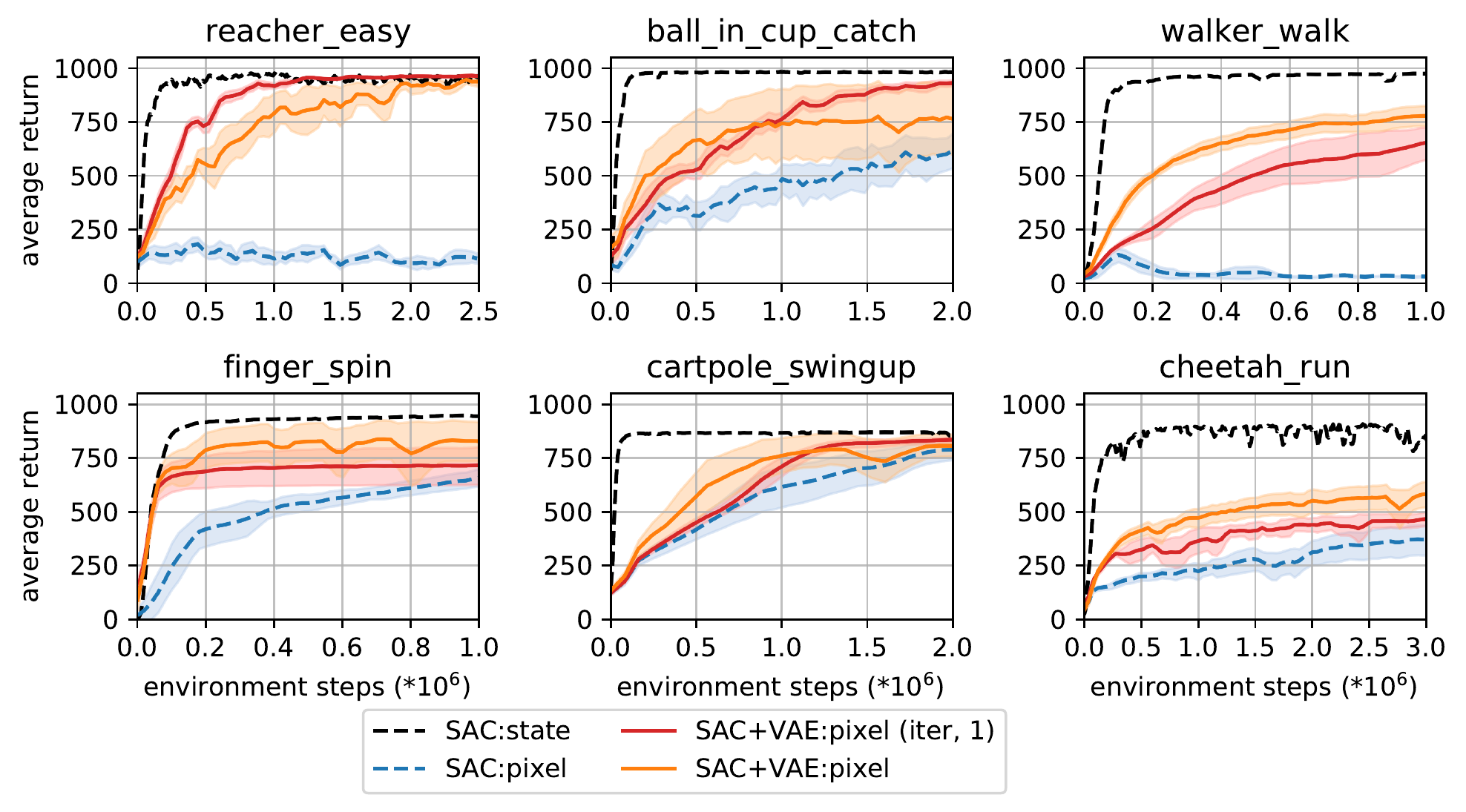}
    \vspace{-5pt}
    \caption{An unsuccessful attempt to propagate gradients from the actor-critic down to the encoder of the $\beta$-VAE to enable joint off-policy training. The learning process of SAC+VAE:pixel exhibits instability together with the subpar performance comparing to the baseline SAC+VAE:pixel (iter, 1), which does not share gradients with the actor-critic.}
    \label{fig:ae_type}
\end{figure}

\newpage 

\section{Stabilizing Joint Representation Learning}
\label{section:dissection:stabilizing_full}
Additional results to the experiments from~\cref{section:dissection:stabilizing} are in~\cref{fig:with_actor_grad_full}.

\begin{figure}[ht!]
    \centering
    \subfloat[Smaller values of $\beta$ reduce stochasticity of a $\beta$-VAE and lead to a better performance.  ]{\includegraphics[width=0.9\linewidth]{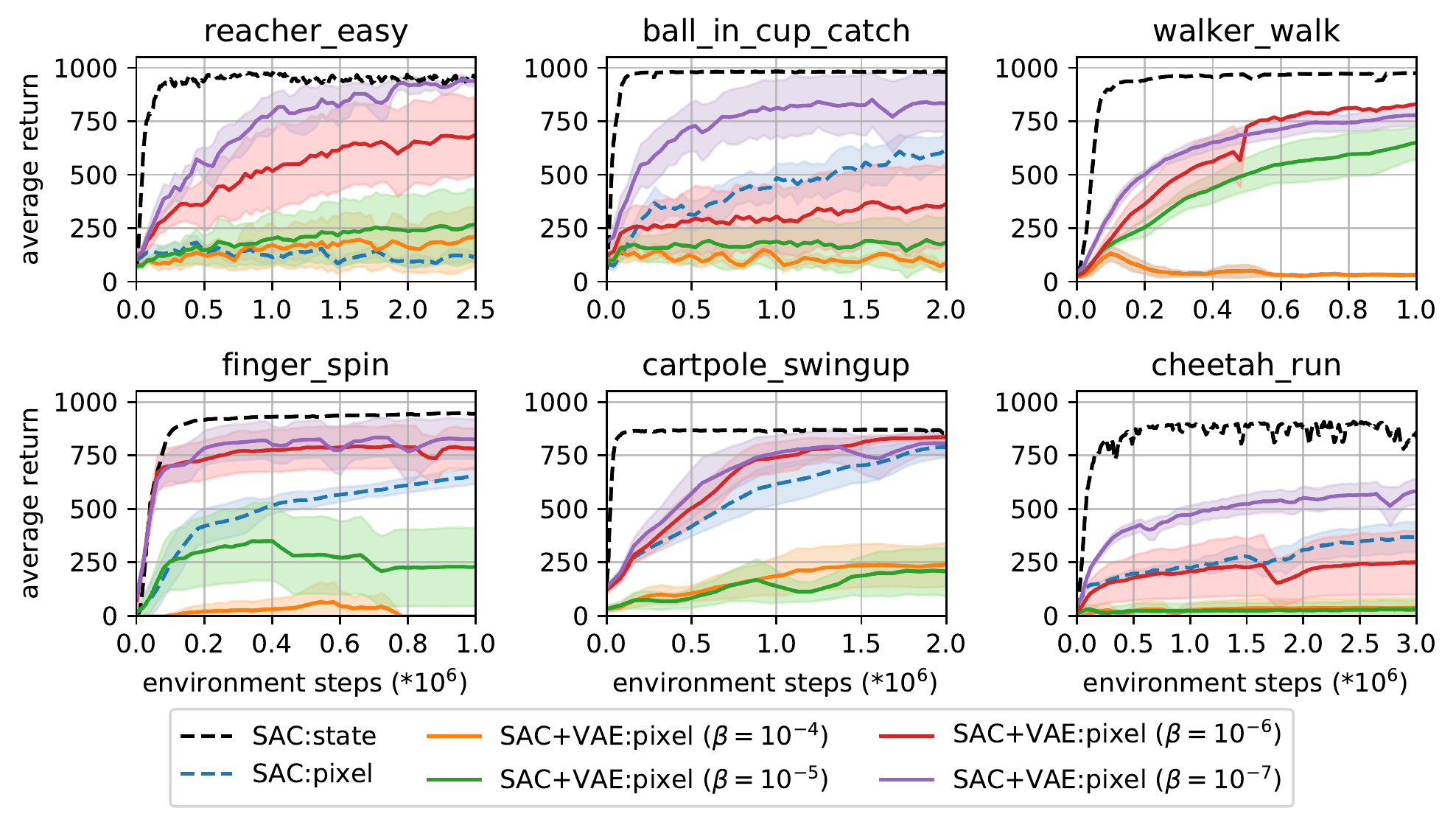}} \\
    \subfloat[Preventing the actor's gradients to update the convolutional encoder helps to improve performance even further.]{\includegraphics[width=0.9\linewidth]{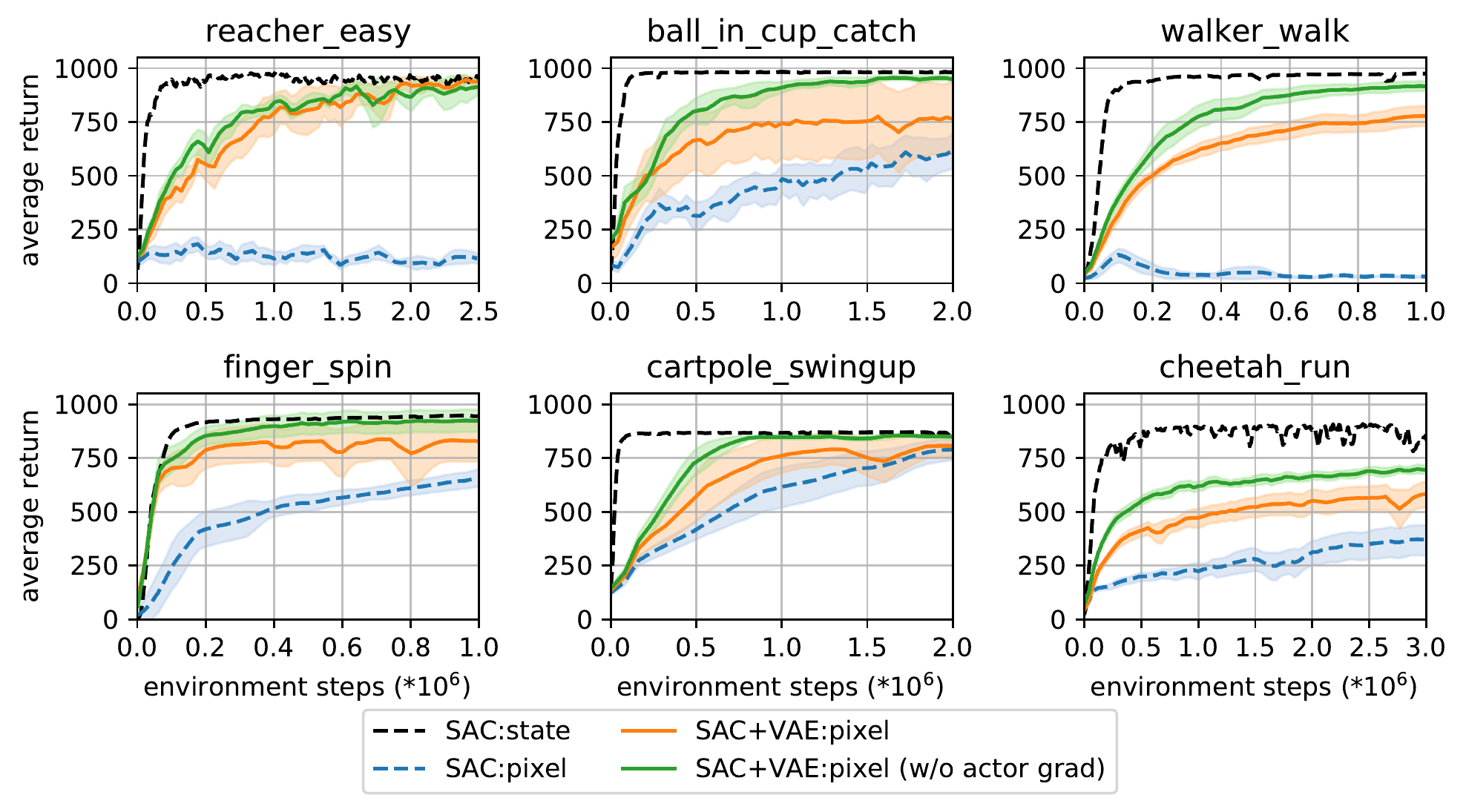}}
        \vspace{-5pt}
    \caption{\small{We identify two reasons for the subpar performance of joint representation learning. (a) The stochastic nature of a $\beta$-VAE, and (b) the non-stationary actor's gradients. }}
    \label{fig:with_actor_grad_full}
\end{figure}

\newpage

\section{Capacity of the Autoencoder}
\label{section:autoencoder_capacity}
We also investigate various autoencoder capacities for the different tasks. Specifically, we measure the impact of changing the capacity of the convolutional trunk of the encoder and corresponding deconvolutional trunk of the decoder. Here, we maintain the shared weights across convolutional layers between the actor and critic, but modify the number of convolutional layers and number of filters per layer in~\cref{fig:rae_arch} across several environments. We find that SAC+AE is robust to various autoencoder capacities, and all architectures tried were capable of extracting the relevant features from pixel space necessary to learn a good policy. We use the same training and evaluation setup as detailed in~\cref{section:hyperparams:training_setup}.
\begin{figure}[hb!]
    \centering
    \includegraphics[width=0.9\linewidth]{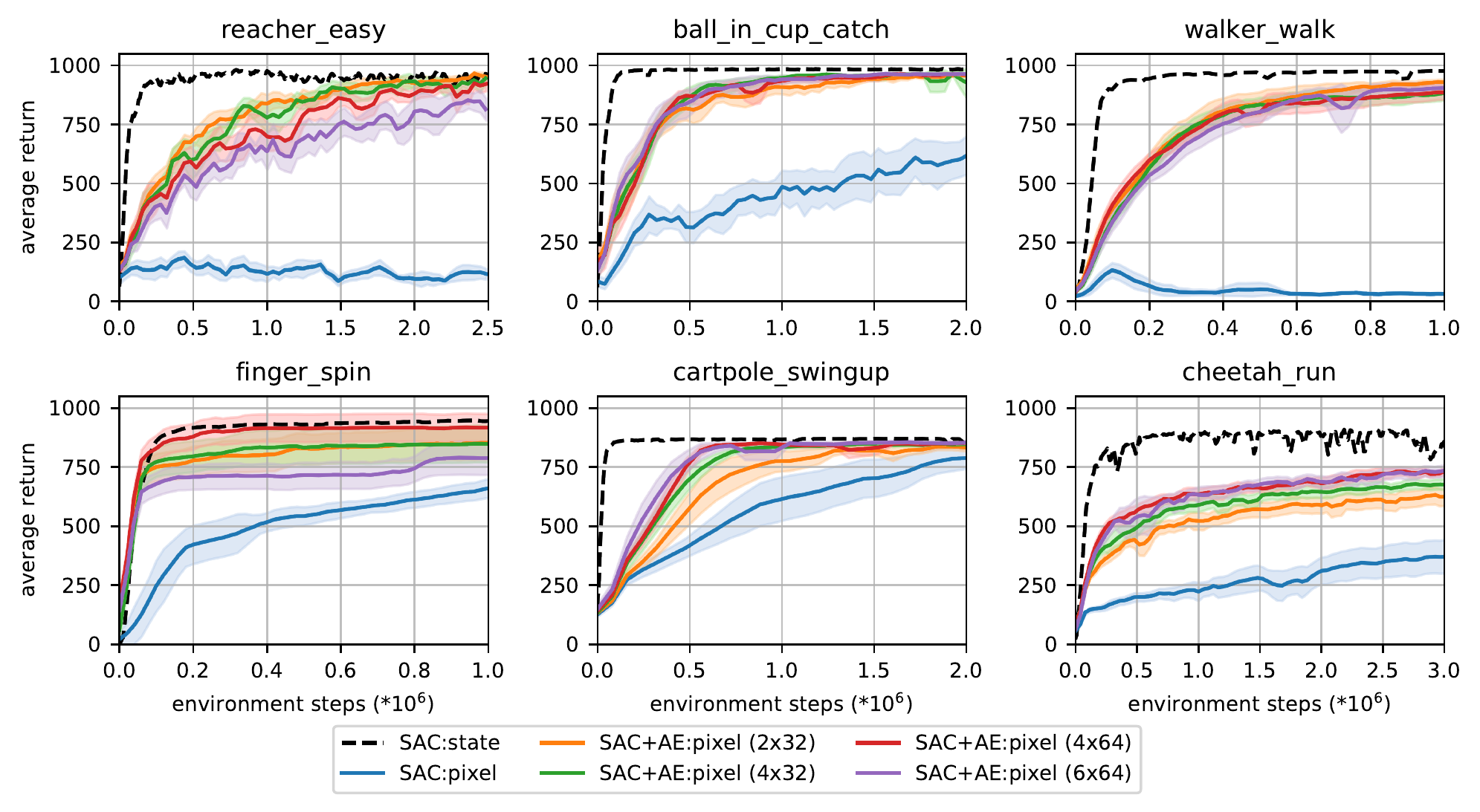}
    \caption{Different autoencoder architectures, where we vary the number of conv layers and the number of output channels in each layer in both the encoder and decoder. For example, $4 \times 32$ specifies an architecture with $4$ conv layers, each outputting $32$ channels. We observe that the difference in capacity has only limited effect on final performance.}
    \label{fig:rae_arch}
\end{figure}

\newpage

\section{Representation Power of the Encoder}
\label{section:ablation:repr_power_full}
Addition results to the experiment in~\cref{section:ablation:repr_power} that demonstrates encoder's power to reconstruct proprioceptive state from image-observations are shown in~\cref{fig:encoder_proprio_rec}.

\begin{figure}[hb!]
    \centering
    \includegraphics[width=0.7\linewidth]{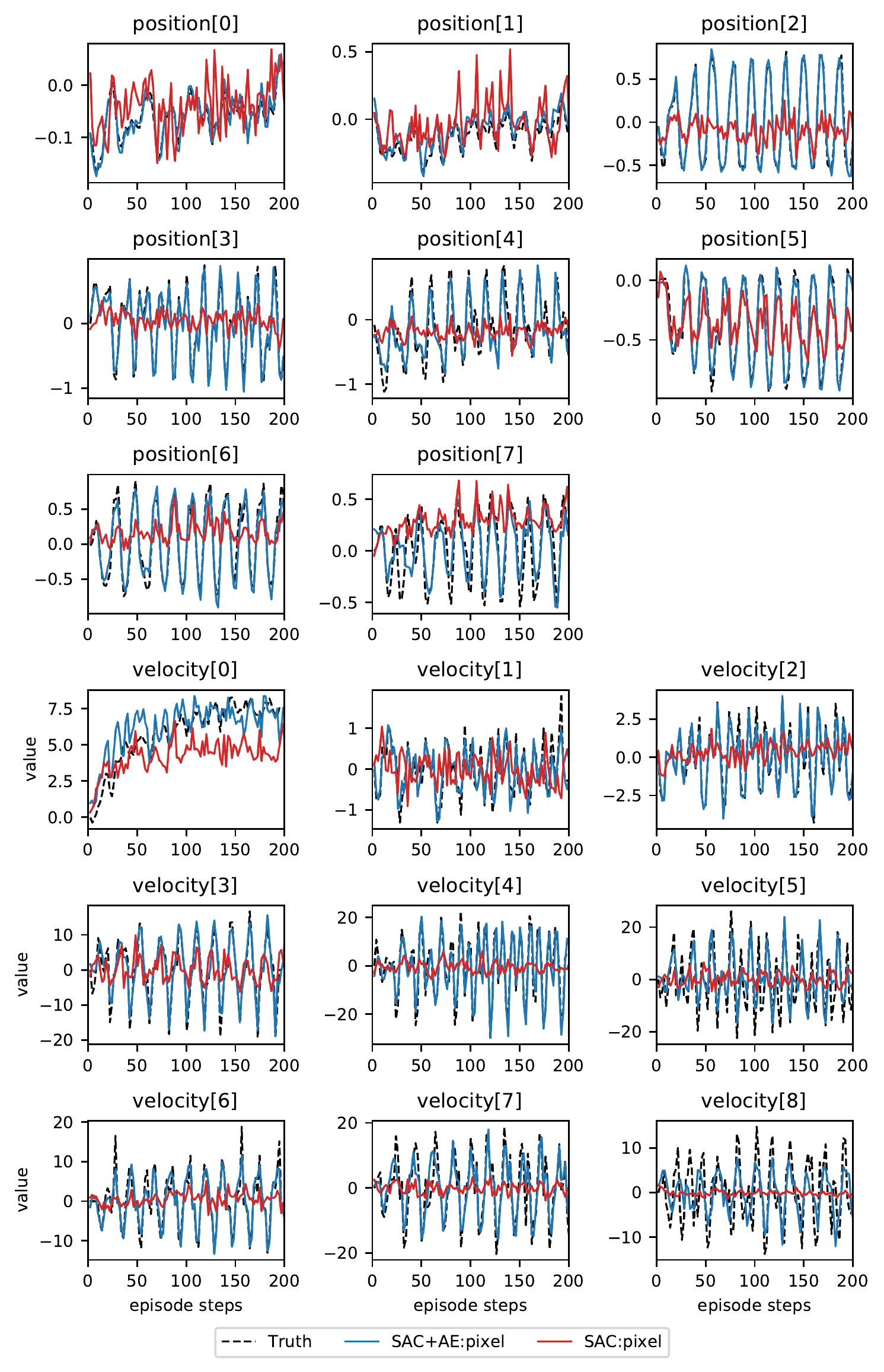}
    \caption{Linear projections of latent representation spaces learned by our method (SAC+AE:pixel) and the baseline (SAC:pixel) onto proprioceptive states. We compare ground truth value of each proprioceptive coordinate against their reconstructions for \texttt{cheetah\_run}, and conclude that our method successfully encodes proprioceptive state information.  The proprioceptive state of \texttt{cheetah\_run} has $8$ position and $9$ velocity coordinates. }
    \label{fig:encoder_proprio_rec}
\end{figure}

\newpage

\section{Decoding to Proprioceptive State}
\label{section:dissection:proprio_reconstruction_full}
Learning from low-dimensional proprioceptive observations achieves better final performance with greater sample efficiency (see~\cref{fig:benchmark_planet} for comparison to pixels baselines), therefore our intuition is to directly use these compact observations as the reconstruction targets to generate an auxiliary signal. Although, this is an unrealistic setup, given that we do not have access to proprioceptive states in practice,  we use it as a tool to understand if such supervision is beneficial for representation learning and therefore can achieve good performance.  We augment the observational encoder $g_\phi$, that maps an image $\rvo_t$ into a latent vector $\rvz_t$, with a state decoder $f_{\theta}$, that restores the corresponding state $\rvs_t$ from the latent vector $\rvz_t$. This leads to an auxililary objective $\E_{\rvo_t, \rvs_t \sim \gD} \big[ \frac{1}{2} ||f_\theta(\rvz_t) - \rvs_t  ||_2^2 \big]$, where $\rvz_t = g_{\phi}(\rvo_t)$. We parametrize the state decoder $f_{\theta}$
as a $3$-layer MLP with $1024$ hidden size and \texttt{ReLU} activations, and train it jointly with the actor-critic network. Such auxiliary supervision helps less than expected, and surprisingly hurts performance in \texttt{ball\_in\_cup\_catch}, as seen in \cref{fig:proprio_reconstruction}. Our intuition is that such low-dimensional supervision is not able to provide the rich reconstruction error needed to fit the high-capacity convolutional encoder $g_{\phi}$. We thus seek for a  denser auxiliary signal and try learning latent representation spaces with pixel reconstructions.

\begin{figure}[hb!]
    \centering

    \includegraphics[width=0.9\linewidth]{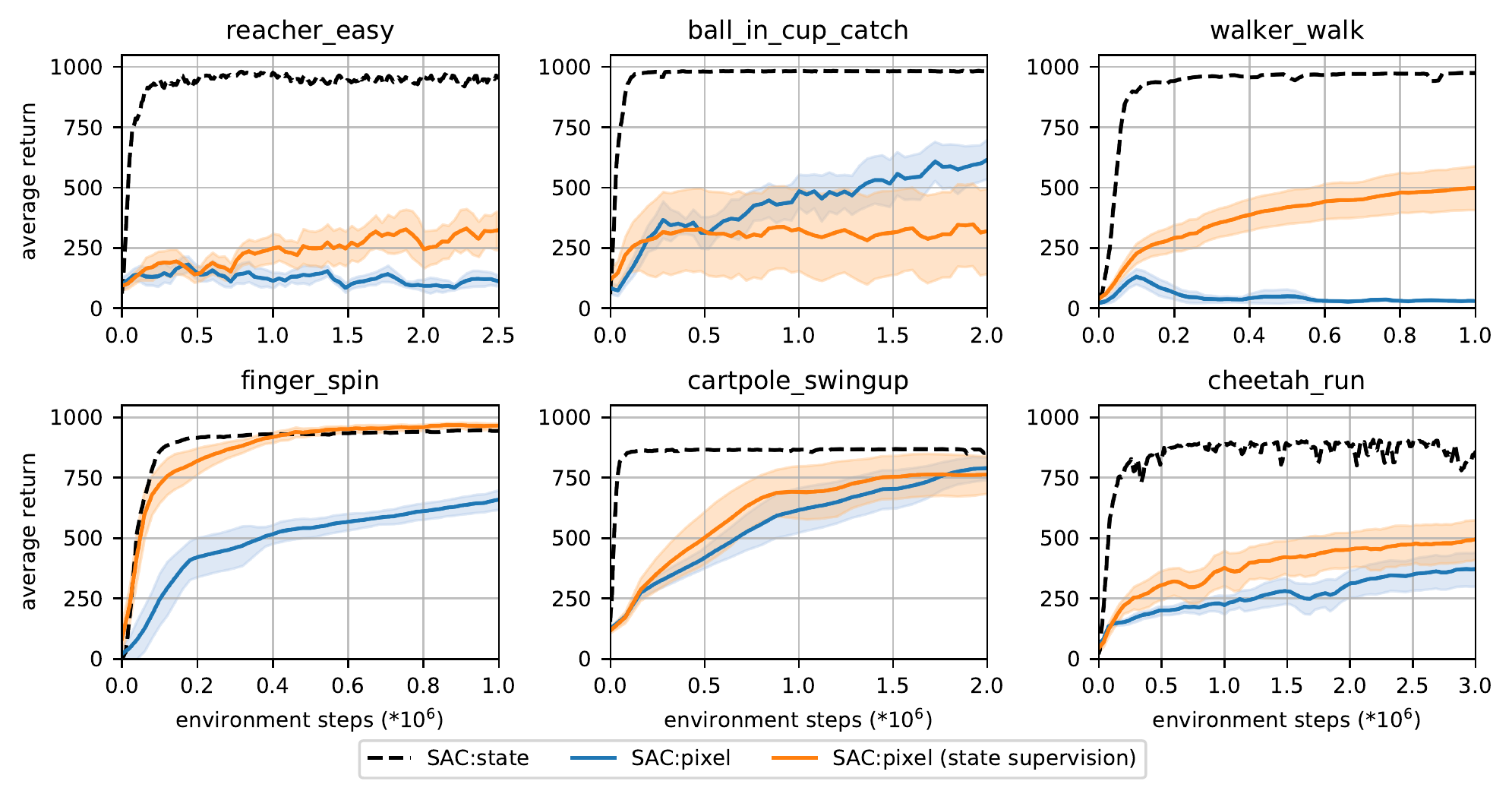}

    \caption{An auxiliary signal is provided by reconstructing a low-dimensional state from the corresponding image observation. Perhaps surprisingly, such \emph{synthetic} supervision doesn't guarantee sufficient signal to fit the high-capacity encoder, which we infer from the suboptimal performance of SAC:pixel (state supervision) compared to SAC:pixel in \texttt{ball\_in\_cup\_catch}.}
    \label{fig:proprio_reconstruction}

\end{figure}
 \newpage

\section{Optimality of Learned Latent Representation}
\label{section:optimality}
We define the optimality of the learned latent representation as the ability of our model to extract and preserve all relevant information from the pixel observations sufficient to learn a good policy. For example, the proprioceptive state representation is clearly better than the pixel representation because we can learn a better policy. However, the differences in performance of SAC:state and SAC+AE:pixel can be attributed not only to the different observation spaces, but also the difference in data collected in the replay buffer. To decouple these attributes and determine how much information loss there is in moving from proprioceptive state to pixel images, we measure final task reward of policies learned from the same fixed replay buffer, where one is trained on proprioceptive states and the other trained on pixel observations. 

We first train a SAC+AE policy until convergence and save the replay buffer that we collected during training. Importantly, in the replay buffer we store both the pixel observations and the corresponding proprioceptive states.
Note that for two policies trained on the fixed replay buffer, we are operating in an off-policy regime, and thus it is possible we won't be able to train a policy that performs as well. 

\begin{figure}[H]
    \centering
    \includegraphics[width=0.9\linewidth]{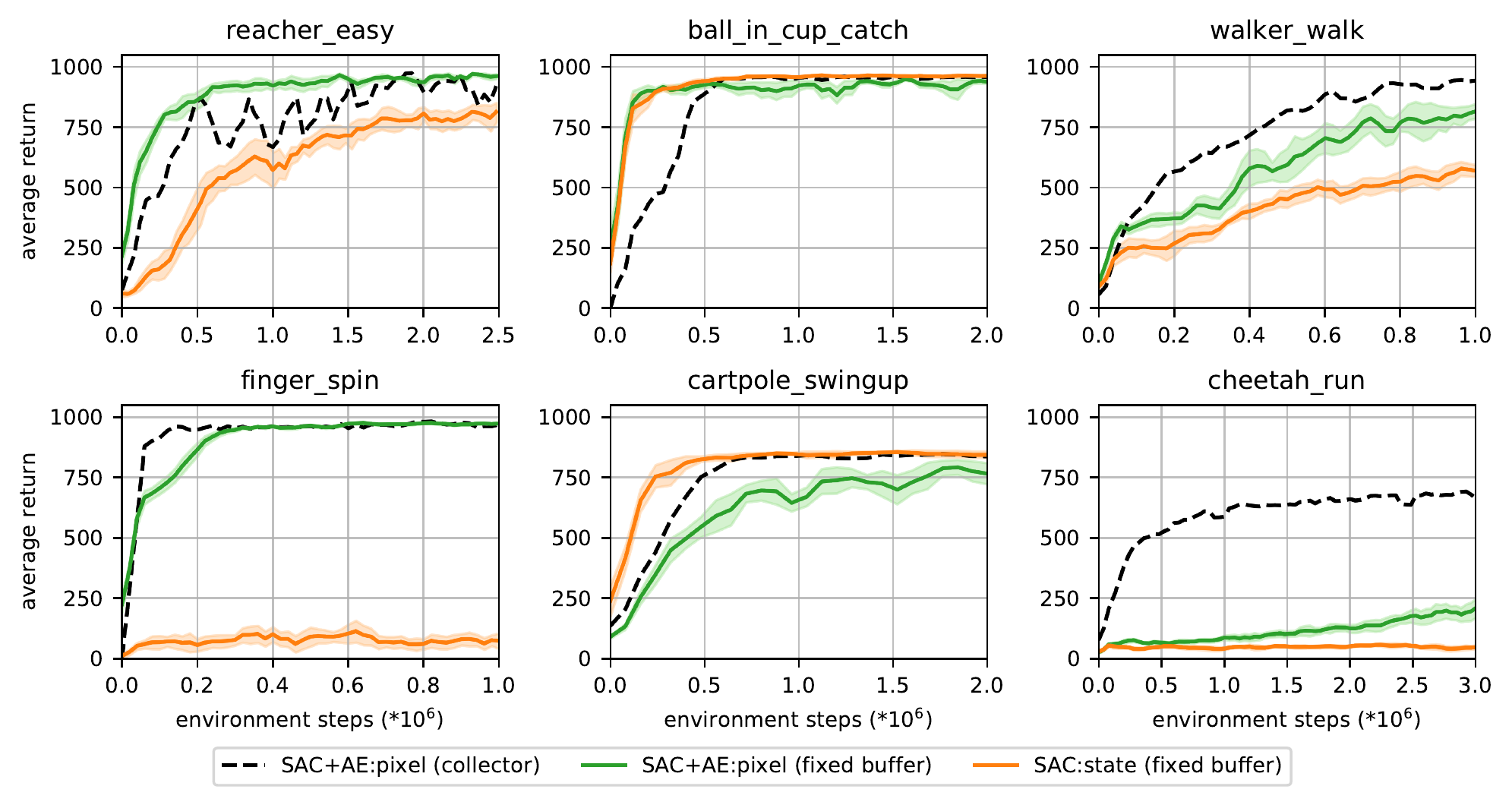}
    \caption{Training curves for the policy used to collect the buffer (SAC+AE:pixel (collector)), and the two policies learned on that buffer using proprioceptive (SAC:state (fixed buffer)) and pixel observations (SAC+AE:pixel (fixed buffer)). We see that our method actually outperforms proprioceptive observations in this setting.}
    \label{fig:optimality}
\end{figure}

In \cref{fig:optimality} we find, surprisingly, that our learned latent representation outperforms proprioceptive state on a fixed buffer. This could be because the data collected in the buffer is by a policy also learned from pixel observations, and is different enough from the policy that would be learned from proprioceptive states that SAC:state underperforms in this setting.

\newpage 

\section{Importance of Action Repeat}
\label{section:action_repeat}

We found that repeating nominal actions several times has a significant effect on learning dynamics and final reward. Prior works~\citep{hafner2018planet,lee2019slac} treat action repeat as a hyper parameter to the learning algorithm, rather than a property of the target environment. Effectively, action repeat decreases the control horizon of the task and makes the control dynamics more stable. Yet, action repeat can also introduce a harmful bias, that prevents the agent from learning an optimal policy due to the injected lag. This tasks a practitioner with a problem of finding an optimal value for the action repeat hyper parameter that stabilizes training without limiting control elasticity too much. 

To get more insights, we perform an ablation study, where we sweep over several choices for action repeat on multiple control tasks and compare acquired results against PlaNet~\citep{hafner2018planet} with the original action repeat setting, which was also tuned per environment. We use the same setup as detailed in~\cref{section:hyperparams:training_setup}. Specifically, we average performance over $10$ random seeds, and reduce the number of training observations inverse proportionally to the action repeat value.  The results are shown in~\cref{fig:action_repeat}. We observe that PlaNet's choice of action repeat is not always optimal for our algorithm. For example, we can significantly improve performance of our agent on the \texttt{ball\_in\_cup\_catch} task if instead of taking the same nominal action four times, as PlaNet suggests, we take it once or twice.  The same is true on a few other environments.

\begin{figure}[H]
    \centering
    \includegraphics[width=0.9\linewidth]{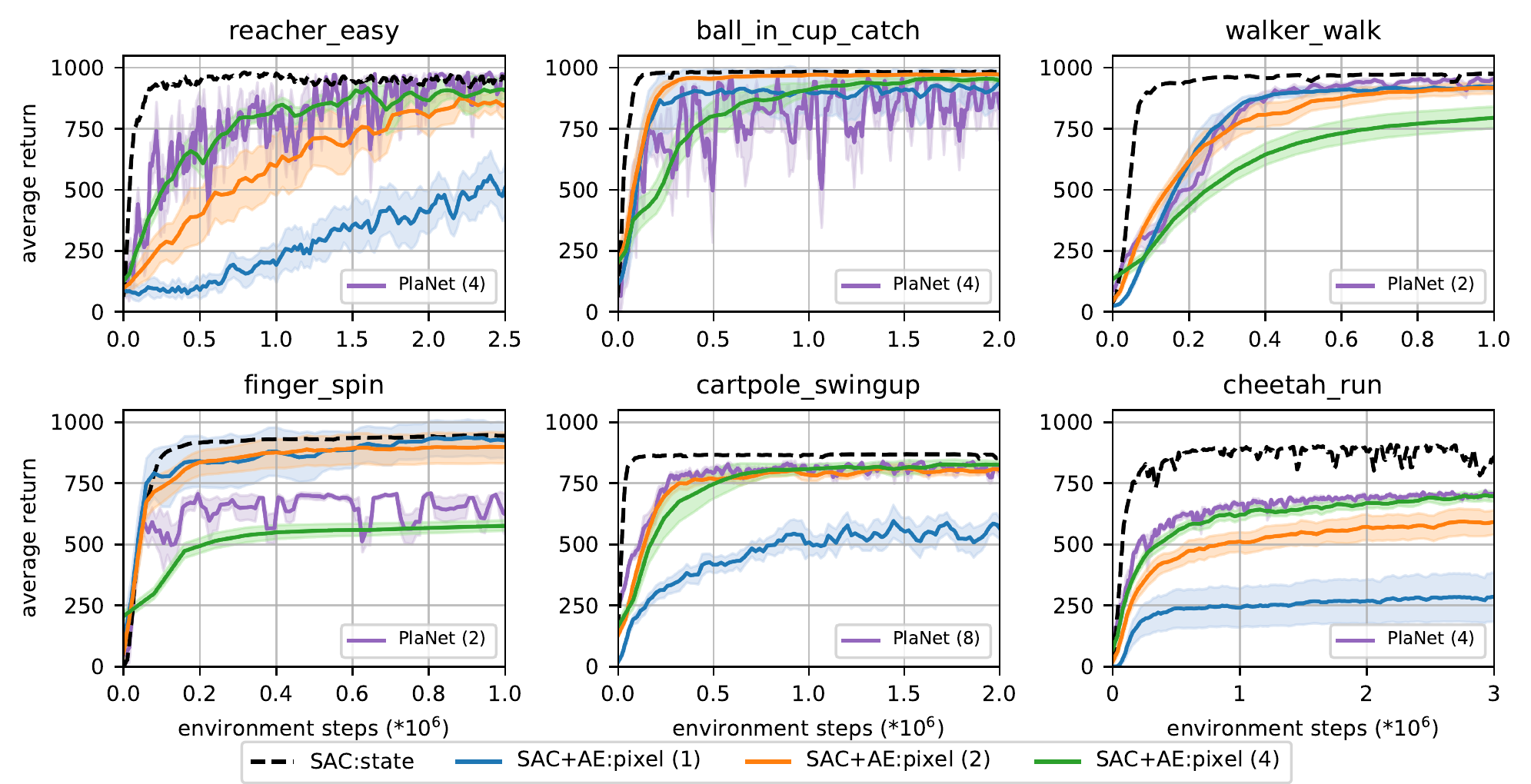}
    \caption{We study the importance of the action repeat hyper parameter on final performance. We evaluate three different settings, where the agent applies a sampled action once (SAC+AE:pixel (1)), twice (SAC+AE:pixel (2)), or four times (SAC+AE:pixel (4)).
    As a reference, we also plot the PlaNet~\citep{hafner2018planet} results  with the original action repeat setting.
    Action repeat has a significant effect on learning.
    Moreover, we note that the PlaNet's choice of hyper parameters is not always optimal for our method (e.g. it is better to apply an action only once on \texttt{walker\_walk}, than taking it twice).}
    \label{fig:action_repeat}
\end{figure}

\fi

\end{document}